\documentclass{article}

\PassOptionsToPackage{numbers, compress}{natbib}
\usepackage[preprint]{neurips_2022}

\usepackage[utf8]{inputenc} % allow utf-8 input
\usepackage[T1]{fontenc}    % use 8-bit T1 fonts
\usepackage{hyperref}       % hyperlinks
\usepackage{url}            % simple URL typesetting
\usepackage{booktabs}       % professional-quality tables
\usepackage{amsfonts}       % blackboard math symbols
\usepackage{amssymb}
\usepackage{nicefrac}       % compact symbols for 1/2, etc.
\usepackage{microtype}      % microtypography
\usepackage{epigraph}
\usepackage{array}
\usepackage[ruled,vlined]{algorithm2e}

\usepackage{graphicx}
\graphicspath{ {./imgs/} }
\usepackage{caption}        % captioning in subfigures
\usepackage{sidecap}        % SCfigure
\usepackage{subcaption}
\usepackage[font=small]{caption}

\usepackage[normalem]{ulem}	% for strike-out text use \sout{text}
\usepackage[toc,page]{appendix}	%for appendices
\usepackage{blindtext}
\usepackage{setspace}
\usepackage{amsmath}
\usepackage{mathtools}
\usepackage{booktabs}
\usepackage{hyperref}
\usepackage{authblk}
\usepackage{lipsum}

\makeatletter
\DeclareRobustCommand\onedot{\futurelet\@let@token\@onedot}
\def\@onedot{\ifx\@let@token.\else.\null\fi\xspace}

\usepackage[dvipsnames]{xcolor}
\definecolor{britishracinggreen}{rgb}{0.0, 0.26, 0.15}
\definecolor{armygreen}{rgb}{0.29, 0.33, 0.13}
\definecolor{bleudefrance}{rgb}{0.19, 0.55, 0.91}
\definecolor{chromeyellow}{rgb}{1.0, 0.65, 0.0}
\definecolor{cadmiumgreen}{rgb}{0.0, 0.42, 0.24}

\usepackage{tikz}
\usetikzlibrary{matrix,chains,positioning,arrows, calc}
\usepackage{float}
\usepackage{amsthm}
\usepackage{multirow}
\usepackage[ruled, vlined ]{algorithm2e}
\usepackage{algorithmic}
\usepackage{wrapfig}
\usepackage{lipsum,booktabs}
\usepackage{mathtools}

\usepackage{listings}
\lstnewenvironment{myverbatim}[1][]{%
  \lstset{
    basicstyle=\small\ttfamily,
    columns=flexible,
    breaklines=true,
    frame=tb,
    #1
  }%
}{}

\usepackage[framemethod=TikZ]{mdframed}
\newcounter{defn}[section] 
\renewcommand{\thedefn}{\arabic{section}.\arabic{defn}}
\newenvironment{defn}[2][]{%
\refstepcounter{defn}%
\ifstrempty{#1}%
{\mdfsetup{%
frametitle={%
\tikz[baseline=(current bounding box.east),outer sep=0pt]
\node[anchor=east,rectangle,fill=blue!20]
{\strut Definition~\thedefn};}}
}%
{\mdfsetup{%
frametitle={%
\tikz[baseline=(current bounding box.east),outer sep=0pt]
\node[anchor=east,rectangle,fill=blue!20]
{\strut Definition~\thedefn:~#1};}}%
}%
\mdfsetup{innertopmargin=10pt,linecolor=blue!20,%
linewidth=2pt,topline=true,%
frametitleaboveskip=\dimexpr-\ht\strutbox\relax
}
\begin{mdframed}[]\relax%
\label{#2}}{\end{mdframed}}
\newcounter{theo}[section] 
\renewcommand{\thetheo}{\arabic{section}.\arabic{theo}}
\newenvironment{theo}[2][]{%
\refstepcounter{theo}%
\ifstrempty{#1}%
{\mdfsetup{%
frametitle={%
\tikz[baseline=(current bounding box.east),outer sep=0pt]
\node[anchor=east,rectangle,fill=blue!20]
{\strut Theorem~\thetheo};}}
}%
{\mdfsetup{%
frametitle={%
\tikz[baseline=(current bounding box.east),outer sep=0pt]
\node[anchor=east,rectangle,fill=blue!20]
{\strut Theorem~\thetheo:~#1};}}%
}%
\mdfsetup{innertopmargin=10pt,linecolor=blue!20,%
linewidth=2pt,topline=true,%
frametitleaboveskip=\dimexpr-\ht\strutbox\relax
}
\begin{mdframed}[]\relax%
\label{#2}}{\end{mdframed}}

\newcounter{lem}[section] 
\renewcommand{\thelem}{\arabic{lem}}
\newenvironment{lem}[2][]{%
\refstepcounter{lem}%
\ifstrempty{#1}%
{\mdfsetup{%
frametitle={%
    \tikz[baseline=(current bounding box.east),outer sep=0pt]
    \node[anchor=east,rectangle,fill=violet!20]
    {\strut Lemma~\thelem};}}
}%
{\mdfsetup{%
frametitle={%
    \tikz[baseline=(current bounding box.east),outer sep=0pt]
    \node[anchor=east,rectangle,fill=violet!20]
    {\strut Lemma~\thelem:~#1};}}%
}%
\mdfsetup{innertopmargin=10pt,linecolor=violet!20,%
linewidth=2pt,topline=true,%
frametitleaboveskip=\dimexpr-\ht\strutbox\relax
}
\begin{mdframed}[]\relax%
\label{#2}}{\end{mdframed}}
\newcounter{prf}[section]
\renewcommand{\theprf}{\arabic{section}.\arabic{prf}}

\theoremstyle{remark}

\theoremstyle{definition}

\title{Reward Learning using Structural Motifs \\
in Inverse Reinforcement Learning}

\author[1,2]{Raeid Saqur}
\author[1,2]{Animesh Garg}
\author[1,2]{Frank Rudzicz}

\affil[1]{\footnotesize [raeidsaqur | garg | frank]@cs.toronto.edu, University of Toronto, ON, Canada}
\affil[2]{\footnotesize Vector Institute}
\newif\ifsubfiles
\subfilesfalse 
\ifsubfiles
    \usepackage{subfiles} % Best loaded last in the preamble 
    \AtBeginDocument{%
    }
    \AtEndDocument{%
        \newpage
        \printbibliography
    }
\fi

\begin{document}
\maketitle

% \todo{Abstract last} \lipsum[1]
\begin{abstract}
The Inverse Reinforcement Learning (\textit{IRL}) problem has seen rapid evolution in the past few years, with important applications in domains like robotics, cognition, and health. In this work, we explore the inefficacy of current IRL methods in learning an agent's reward function from expert trajectories depicting long-horizon, complex sequential tasks. We hypothesize that imbuing IRL models with structural motifs capturing underlying tasks can enable and enhance their performance. Subsequently, we propose a novel IRL method, SMIRL, that first learns the (approximate) structure of a task as a finite-state-automaton (FSA), then uses the structural motif to solve the IRL problem. We test our model on both discrete grid world and high-dimensional continuous domain environments. We empirically show that our proposed approach successfully learns all four complex tasks, where two foundational IRL baselines fail. Our model also outperforms the baselines in sample efficiency on a simpler toy task. We further show promising test results in a modified continuous domain on tasks with compositional reward functions.  
\end{abstract}

%=====================
%   section.tex files
%=====================
% ============================================================================================
\section{Introduction}\label{intro}
% ============================================================================================

\begin{wrapfigure}[18]{R}{0.40\textwidth} 
    \centering
    \hspace{-6pt}
    \vspace{-6pt}
    %\vspace{-\baselineskip}
    \includegraphics[width=0.38\textwidth]{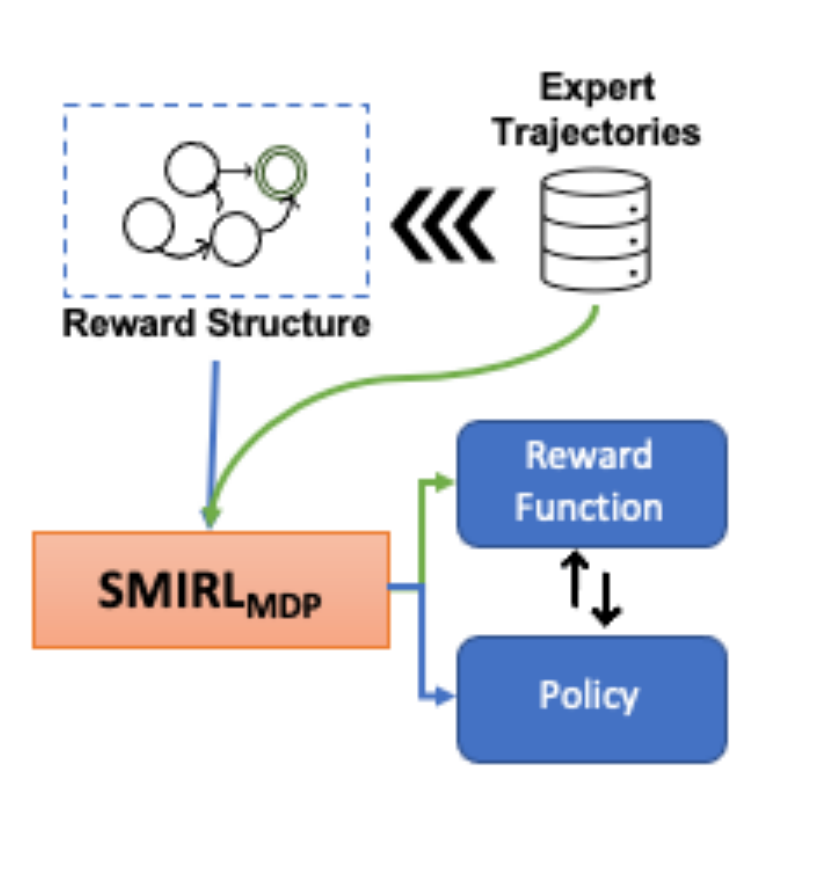}
    % \vfill 
    \vspace{4pt}
    \caption{A high-level schematic diagram of our proposed IRL model: SMIRL}
    \label{fig:model_diagram}
    %\vspace{-6pt}
\end{wrapfigure}

Inverse Reinforcement Learning (IRL)~\cite{ng2000algorithms} has evolved considerably since first introduced by Russell~\cite{russell1998learning}. It has been applied in various domains and contexts, including robotics~\cite{argall2009survey}, cognition~\cite{baker2009action}, and health~\cite{asoh2013application}. The IRL problem entails inferring the underlying reward structure, that incentivizes an agent's behaviour, based on observations, and a model of the environment in which the agent acts. The early works in IRL entailed representing the reward function as a weighted linear combination of handcrafted features~\cite{abbeel2004apprenticeship}. Essentially, a strategy of matching \textit{feature expectations} between an observed policy and an agent's behaviour~\cite{arora2021survey}. However, such approaches have inherent weaknesses in delineating trajectories from sub-optimal policies (e.g., consider the degenerate case of all zeroes). More recent approaches~\cite{maxentIRL2008, wulfmeier2015maximum, f-irl2020, avril2021} circumvent this weakness using max-entropy~\cite{jaynes1957information} based principled approaches that consider a distribution over all possible behaviour trajectories and favour the trajectories with less ambiguity. 

While these approaches show demonstrable success over tasks formulated as MDP problems~\cite{2022survey,arora2021survey}, in this work we explore their efficacy in long-horizon, temporally extended, complex sequential tasks, where the rewards received by an agent are \textbf{not necessarily Markovian} with respect to the state space. 
To this accord, we conduct experiments first on a simple 2D discrete environment (S.\ref{sec:experiments-gridworld}), followed by a high-dimensional continuous domain environment (S.\ref{sec:experiments-reacher}) using tasks with varying complexity between observations and underlying state reward. We mimic causal structure among states' propagation with logical conditions~\cite{vardi1996automata, giacomo1999automata}. To elucidate, a simple MDP problem of moving to a specific location on a map (\textit{Task 0}) [Fig.~\ref{fig:office-world-grid}] were modified with more challenging long-horizon tasks (\textit{Tasks 1-4})~[Fig.~\ref{fig:rm-task4}], like patrolling a set of locations in conditional (e.g., sequential) orders to receive rewards. We found that while both baselines succeeded in the simple task, they completely failed on the harder tasks.

We hypothesize that imbuing IRL models with a \textit{structural motif} of the underlying complex task can enable the models to learn the reward functions even in such long-horizon complex causal scenarios. Based on our hypothesis, we propose a novel IRL model (SMIRL) (Fig.~\ref{fig:model_diagram}) that learns a finite-state-automaton (FSA) based reward structure as motif, then uses the structural motif in solving the IRL problem. 

% -- Reward Machine (RM)~\cite{qrm2018, lrm2020} -- then uses the RM in solving the IRL problem. 
% Main technical contribs.
We empirically show that our proposed model is able to successfully learn four complex sequential tasks in the \textit{ Office Grid} environment where the other baselines using foundational IRL approaches (Apprenticeship~\cite{abbeel2004apprenticeship}, MaxEnt~\cite{maxentIRL2008, wulfmeier2015maximum}) categorically fail.

% on established baselines from the IRL literature~\cite{abbeel2004apprenticeship, wulfmeier2015maximum},
% To this accord, we conducted  experiments in the \textit{Office Grid World} environment, where two popular baselines from the IRL literature~\cite{abbeel2004apprenticeship, wulfmeier2015maximum} were first tasked with a simple MDP problem of moving to a specific location on a map (\textit{Task 0}), and then with more challenging long-horizon tasks (\textit{Tasks 1-4}), like patrolling a set of locations in sequential order to receive rewards. 

% The agent's reward functions for tasks in this environment can be expressed using simple automaton (FSA) based representation like `Reward Machines' [RM]~\cite{}.

% As our IRL baselines, we try out a few popular and SOTA IRL approaches to infer the reward function. We then compare the performance of our approach of first trying to infer the structural motif in the reward function using the LRM~\cite{} approach.

% \paragraph{Discussion} Classical IRL approaches work with the assumption that the belief states and observations are Markovian w.r.t the reward function, however, for cases (like the e.g. used by us) where this assumption doesn't hold, IRL approaches are not effective.

% =====================================================================
\section{Preliminaries and Background}\label{sec:preliminaries}
% =====================================================================
\paragraph{Markov Decision Process (MDP)} A MDP is defined by a tuple $(S, A, T, R, \gamma, p_0)$ where $S$ is a set of states (state space), $A$ is a set of actions (action space), $T: S\times A \to \Pi(S)$ is the transition function, $R: S \to \mathbb{R}$ is the reward function, $\gamma \in [0, 1]$ is the discount factor, and $p_0: S \to [0, 1]$ is the distribution over initial states. A policy over a MDP is a function $\pi: S \to \Pi(A)$, and is optimal if it maximizes the expected discounted sum of rewards.
% $L = \mathbb{E}_{\pi, T}\left(\sum_{s_i \in \tau} \gamma^i R(s_i)\right)$, where $\tau = (s_0, a_0, \dots, s_T)$ is a trajectory.

\paragraph{Partially Observable Markov Decision Process (POMDP)} A POMDP is a generalisation of a MDP defined by the tuple $(S, A, T, O, \omega, R, \gamma, p_0)$ where $O$ is a set of observations and $\omega: S \to \Pi(O)$ is the observation function. An agent in a POMDP thus only receives an observation (i.e., partial information about the state) rather than the actual state of the environment. Therefore, policies on POMDP act based on the history of observations received and actions taken at timestep $t$. Since using the complete history is impractical, many algorithms instead use a belief $b: O \to \Pi(S)$, which is a probability distribution over possible states updated at each timestep. The belief update after taking the action $a\in A$ and receiving observation $o \in O$ is done through the following equation:
\begin{align}
    \label{eqn:belief}
    b_{o}^{a}\left(s^{\prime}\right)=P\left(s^{\prime} \mid b, a, o\right)=\frac{\omega\left(s^{\prime}, o\right) \sum_{s} T\left(s, a, s^{\prime}\right) b(s)}{P(o \mid b, a)} \;\; \forall s' \in S,
\end{align}
where $P(o \mid b, a)=\sum_{s^{\prime}} \omega\left(s^{\prime}, o\right) \sum_{s} T\left(s, a, s^{\prime}\right) b(s)$.

%% Rest in 03_related w/o section heading
% =====================================================================
% \section{Related Work}
% =====================================================================
\label{sec:related-work}
\paragraph{Inverse Reinforcement Learning (IRL)}
Given some demonstrations $\mathcal{D}=\left\{\tau_{1}, \tau_{2}, \ldots, \tau_{N}\right\}$ generated by an expert policy $\pi_E$, the goal of inverse reinforcement learning \cite{ng2000algorithms, abbeel2004apprenticeship, arora2021survey, konar2021learning} is to learn a reward function that can explain the behaviour of the expert policy. Unlike \textit{imitation learning} \cite{hussein2017imitation}, learning reward functions can provide better generalizations of the agent when it comes to a new environment compared to only mimicking expert behaviours. 
% Maybe add a sentence on the fact that the IRL problem is ill-posed

\paragraph{IRL Approaches} 
We use two popular and foundational approaches from the dense IRL literature~\cite{arora2021survey} as our baselines. `\textit{Apprenticeship Learning}'~\cite{abbeel2004apprenticeship} is a seminal work and still relevant today~\cite{escontrela2022adversarial}. This approach entails calculating feature expectations for the states of the expert  and the apprentice (learner) trajectories separately, then optimizing the reward function parameters using max-margin, or quadratic programming (QP). The learner's feature expectations change along the learning process. An iterative procedure is then usually used to update the learner's policy and the reward function for each state representation, until the algorithm converges.

Unfortunately, this approach of matching feature counts between expert and learner is ambiguous. Sub-optimal policies can lead to the same feature counts (including the degenerate cases, e.g., all zeroes). The `\textit{Maximum Entropy} IRL'~\cite{maxentIRL2008} approach deals with this ambiguity in a principled way using the maximum entropy principle~\cite{jaynes1957information}. Broadly, this approach considers a distribution over all possible behaviour trajectories, and disfavours ambiguous policies with high entropy, choosing the least ambiguous policy that does not exhibit additional information beyond matching feature expectations. The resulting distribution over trajectories ($\tau_i$) for deterministic MDPs is parameterized by reward weights $\theta$, such that, trajectories with equivalent rewards have equal probabilities, and trajectories with higher rewards are preferred exponentially more (Eq.~\ref{eqn:max-ent-prob}):
\begin{equation}\label{eqn:max-ent-prob}
    P(\tau_i \mid \theta) = \frac{1}{Z(\theta)}e^{(\theta^T\mathbf{f}_{\tau_i})}
\end{equation}
where $Z(\theta)$ is the partition function, $\tau$ is the trajectory, and $f_{\tau_i}$ is the feature representation.

% \todo{Related work: add or discuss , GAIL~\cite{gail2016}, f-IRL~\cite{f-irl2020}, AVRIL~\cite{avril2021}}
% Reptitive:
% i.e.,  
% \begin{equation}
%     \mu (\pi_e) = \mathbb{E}_{\tau}\left[ \sum_{t=0}^{\infty} \gamma^t s_t^h \biggr{|} \pi_e\right], \mu (\pi_l) = \mathbb{E}_{\tau}\left[ \sum_{t=0}^{\infty} \gamma^t s_t^h \biggr{|} \pi_l\right] 
% \end{equation}
% where $\pi_l$ denotes the learner's policy, and 
% $\pi_e$ denotes the expert's policy.
% With those feature expectations, one can optimize to get the reward function parameters by doing max-margin optimization:
% \begin{equation}
% \begin{split}
%   QP(\mathbf{t}) = max & \quad t\\
%     s.t. & \quad [\mu (\pi_e) - \mu (\pi_l)]^{\intercal} \mathbf{w} \geqslant t\\
%     &\quad ||\mathbf{w}||\leqslant 1
% \end{split}
% \end{equation}
% Approaches include~\cite{abbeel2004apprenticeship, konar2021learning}. 
\paragraph{IRL with Partial Observability}
Choi \& Kim \cite{choi2011inverse}, extending their earlier work \cite{choi2009}, tackled the problem of IRL in partially observable domains (POMDP). The authors introduced algorithms in two settings -- one where the expert policy is explicitly given, and one where only the expert trajectories are available. The latter is relevant to our work here. However, unlike our algorithm, introduced in Section \ref{sec:model}, the authors assumed full access to the environment transition probabilities in order to compute the trajectory of beliefs using Eq.~\ref{eqn:belief}. Once such a trajectory is obtained, usual IRL algorithms can be applied to the belief-state MDP.

\paragraph{Reward Machine (RM)} 
RMs \cite{qrm2018,icarte2020reward} are a type of finite state automaton (FSA) that can be used to define the reward function inside a MDP. Given a set of propositional symbols $\mathcal{P}$ (which are linked to the environment states through a labelling function $L: S \to 2^{\mathcal{P}}$), a reward machine (RM) is defined by a tuple $(U, u_0, \delta_u, \delta_r)$, where $U$ is a finite set of states, $u_0 \in U$ is the initial state, $\delta_u : U \times 2^{\mathcal{P}} \to U$ is the state-transition function and $\delta_r : U \to [S \to \mathbb{R}]$ is the state-reward function. In other words, the RM's state is updated at each time-step through $\delta_u$ and the reward that the agent gets from the environment is defined by the reward function output by $\delta_r$. 

It is important to note that RMs can express temporally extended tasks, which are therefore non-Markovian. However, we can define an MDP over the cross product $S \times U$, which keeps its Markovian property.
Expressing the reward function in such a way allows an RL agent to decompose a task into structured sub-problems and learn them effectively. Icarte {\em et al.}~\cite{qrm2018} introduced the Q-learning for RM (QRM), an algorithm based on Q-learning that simultaneously learns a separate policy, $\Tilde{q_u}$, for each state ($u \in U$) in the RM.

Another interesting use of RMs is introduced by Icarte {\em et al.}~\cite{lrm2020}. Through discrete optimization, a RM can be learned from a set of trajectories (\textbf{LRM}). The learned reward can then be used to ease the learning in the environment, especially in POMDPs. The authors notably showed that if a POMDP has a finite belief space, then there exists a RM for an adequate labelling function $L$ such that the POMDP extended with the RM is Markovian with respect to the observations (i.e., the entire history can be explained by only the last observation and the current RM state).

% \todo{The algorithm implemented in the baseline is also referred to as Max-Margin Prediction (MMP) with feature expectation in the litterature. Zhibo maybe you should mention that?}
% \todo{Adversarial learning approaches to IRL }
% See these works~\cite{ho2016generative, fu2017learning, finn2016guided}, and most recently VDB or VAIRL~\cite{peng2018variational}. Imitation learning (IL) (behaviour cloning) methods are orthogonal as no explicit reward functions are learned unlike our (IRL) problem formulation.

% ===========================================================================================
\section{\underline{S}tructural \underline{M}otif-Conditioned \underline{IRL} (SMIRL)} \label{sec:model}
% \section{Reward Machine conditioned Max. Entropy (LRM-MaxEnt)} \label{sec:model}
% \section{Reward Augmented Expectation in Decision Networks} \label{sec:model}
% ===========================================================================================

Given a POMDP $\mathcal{M}=(S, A, T, O, \omega, R, \gamma, p_0)$, we consider the IRL problem of learning a reward function $R_{\theta}$ given a set of expert trajectories $\mathcal{D} = \{\tau_i\}$ where $\tau_i = (o_0, a_0, o_1, a_1, ..., a_T, o_T)$, such that an optimal policy for $\mathcal{M}_{\theta}=(S, A, T, O, \omega, R_{\theta}, \gamma, p_0)$ is also optimal for $\mathcal{M}$. We assume that we have access to $\mathcal{M}\setminus\{R\}$ when learning $R_\theta$. Moreover, since we want to use reward machines, we assume that we have access to a relevant set of propositions $\mathcal{P}$ and the corresponding labelling function $L: O \to 2^{\mathcal{P}}$. 
% We will discuss the limitations of this assumption as well as what we mean by relevant in Section \ref{sec:lims}. \todo{talk about it}

The main idea behind our algorithm (Fig.~\ref{fig:model_diagram}) is to first learn a FSA structure $(U, u_0,  \delta_u)$ on the POMDP using the trajectories in $\mathcal{D}$, and then use MaxEnt IRL \cite{maxentIRL2008} on the resulting extended POMDP to learn the RM's state-reward function $\delta_{r, \theta}$, where $\theta$ represents the learnable parameters of the function.

\textbf{Learning the Reward Machine}. In order to learn the RM's structure, we build on the method described in \cite{lrm2020}, where \textit{Tabu search}~\cite{glover1998tabu} is used to compute an RM that minimize the following objective:
\begin{align}
    \underset{\left\langle U, u_{0}, \delta_{u}\right\rangle}{\operatorname{argmin}} \sum_{\tau \in \mathcal{D}} \sum_{o_t, a_t \in \tau} \log \left(\left|N_{x_{\tau, t}, L\left(o_t\right)}\right|\right),
\end{align}
where $x_{\tau, t}$ refers to the state of the RM at timestep $t$ when rolling out trajectory $\tau$, and $N_{u, l} \in 2^{2^{\mathcal{P}}}$ is the set of all the next abstract observations (i.e., outputs of the labelling function $L$) seen from the RM state $u$ and the abstract observations $l$. In other
words,  $l' \in N_{u, l}$ iff $u=x_{\tau, t}$, $l=L(o_t)$ and $l'=L(o_{t+1})$ for some trace $\tau$ at timestep $t$. The search is further constrained by imposing a limit $u_{max}$ to the number of states of the RM.

Once the FSA structure is learned, we can update the trajectories in $\mathcal{D}$ with the corresponding RM states, i.e., $\mathcal{D}' = \{\tau_i'\}$ where $\tau_i' = (o_0, u_0, a_0, o_1, u_1, a_1, ..., o_T, u_T)$. We assume for the rest of this section that the learned RM is `perfect' (using the vocabulary introduced in \cite{lrm2020}), meaning that there is a MDP defined on the state space $U \times O$, denoted $\mathcal{M}_{RM}$, such that any optimal policy for $\mathcal{M}_{RM}$ is optimal for $\mathcal{M}$. We thus can learn a reward function on $\mathcal{M}_{RM}$ instead of $\mathcal{M}$.

\begin{algorithm}[t]
\SetAlgoLined
\KwIn{Set of expert trajectories $\mathcal{D}$}
 Learn a FSA structure $(U, u_0, \delta_u)$\;
 Update the trajectories in $\mathcal{D}$ with the corresponding RM states\;
 \For{$(u, o) \in U \times O$}{
  $\mu_{\mathcal{D}'}(u,o) \leftarrow \sum_{\tau' \in \mathcal{D}'} \sum_{o_{t}, u_{t}, a_{t} \in \tau'} \mathbf{1}_{u=u_{t}}\mathbf{1}_{o=o_{t}}$
 }
 Initialize $\theta$\;
 \For{$m=1$ to $M$}{
  Compute $\pi_\theta$ optimal policy over $\mathcal{M}_{RM, \theta}$\;
  \For{$(u, o) \in U \times O$}{
   $\mu_{\pi_\theta}(u,o) \leftarrow \mathbb{E}_{\tau \sim \pi_\theta}\left( \sum_{o_{t}, u_{t} \in \tau} \mathbf{1}_{u=u_{t}}\mathbf{1}_{o=o_{t}}\right)$
 }
  Compute $\frac{\partial L}{\partial \theta} = \sum_{u,o \in U \times O} \left( \mu_{\mathcal{D}'}(u, o) - \mu_{\pi_\theta}(u, o) \right) \frac{\partial \delta_{r, \theta}(u, o)}{\partial \theta}$\;
  Update $\theta$\;
 }
 \caption{SMIRL}
 \label{alg:LRM-MaxEntIRL}
\end{algorithm}

\textbf{Learning the Reward Function}.  As $\mathcal{D}'$ can be seen as a set of trajectories in $\mathcal{M}_{RM}$, we can now use IRL algorithms that apply to MDPs such as MaxEnt IRL to learn the reward function, that we now denote $\delta_{r, \theta}: U \times O \to \mathbb{R}$. MaxEnt IRL assumes a maximum entropy constraint on the distribution of possible trajectories. In our case, this translates to:
\begin{align}
    P_\theta(\tau') \propto \exp\left(\sum_{o_t, u_t, a_t \in \tau'} \delta_{r, \theta}(u_t, o_t)\right)
\end{align}
An optimal reward function can then be learned by maximizing the log-likelihood $L$ of the expert trajectories in $\mathcal{D}'$:
\begin{align}
    \theta^* = \underset{\theta}{\operatorname{argmax}} \, L(\theta) = \underset{\theta}{\operatorname{argmax}} \, \sum_{\tau' \in \mathcal{D}'} \log P_\theta(\tau').
\end{align}
As shown in \cite{medirl2016}, the above $\theta^*$ can be computed by gradient ascent using:
\begin{align}
    \frac{\partial L}{\partial \theta} = \sum_{u,o \in U \times O} \left( \mu_{\mathcal{D}'}(u, o) - \mu_{\pi_\theta}(u, o) \right) \frac{\partial \delta_{r, \theta}(u, o)}{\partial \theta},
\end{align}
where $\mu_{\mathcal{D}'}$ is the \textit{state visitation frequency} (SVF) of the expert demonstrations and $\mu_{\pi_\theta}$ is the SVF of the policy optimal with respect to $\mathcal{M}_{RM, \theta}$ (i.e., $\mathcal{M}_{RM}$ with the learned reward function $\delta_{r, \theta}$). 
% \todo{Do I put the formula for SVF? a bit redundant with algorithm box so depends if we're tight on space}

It is common practice \cite{arora2021survey,konar2021learning} in IRL to apply the reward function to handcrafted features $\phi: S \to \mathbb{R}^N$ instead of directly using the raw state. In our experiments (Sec. \ref{sec:experiments}), we use the labelling function $L: O \to 2^{\mathcal{P}}$ as the feature function. The reward at state $u, o \in U \times O$ is therefore $\delta_r(u, L(o))$. We further assume that the reward function is linear with regards to $L(o)$ (with $L(o) \in \{0, 1\}^{|\mathcal{P}|}$), i.e.,
\begin{align}
    \forall u \in U, \; \exists \, \mathbf{w}_u \in \mathbb{R}^{|\mathcal{P}|} \;\;\text{s.t.} \;\; \delta_r(u, L(o)) = \mathbf{w}_u^\top L(o)
\end{align}

%=======================%

% ================================================================================
%%%%%%%%%%%%%%%%%%%%%%%%%%%%%%%%%%%%%%%%%%%%%%%%%%%%%%%%%%%%%%%%%%%%%%%%%%%%%%%%
\section{Experiments}\label{sec:experiments}
%%%%%%%%%%%%%%%%%%%%%%%%%%%%%%%%%%%%%%%%%%%%%%%%%%%%%%%%%%%%%%%%%%%%%%%%%%%%%%%%
% ================================================================================
\subsection{Baselines}
\textbf{Apprenticeship-IRL} For the discrete \textit{Office Gridworld} experiments, we use the classical \textit{Apprenticeship} IRL model~\cite{abbeel2004apprenticeship} for both inferring reward function and learning an optimal agent from given expert trajectories, $\mathcal{D}_e$. The initial step entails learning the environment's high-level feature expectations using Monte Carlo estimate on the $\mathcal{D}_e$ samples for the expert and simulated trajectories by an initialized policy, $\pi_l$ (e.g., a Q-table) for the learner. 
\begin{equation}\label{eq:feat-exp}
    \mu (\pi) = \mathbb{E}_{\tau}\left[ \sum_{t=0}^{\infty} \gamma^t s_t^h \biggr{|} \pi\right] \approx \sum_{i=1}^n \sum_{t=0}^{\infty} \gamma^t s_t^h \in \mathbb{R}^9
\end{equation}
where $s_t^h$ represents the high-level features at time step $t$. The outer layer of summation is over different trajectories, and the inner layer of summation is over time steps within one trajectory. 

As per the IRL definition, the goal is to learn a weight $\mathbf{w}$  such that the reward can be modelled as a weighted linear function of the high-level state features $R(s_t) = \mathbf{w}^{\intercal} s_t^h$. Quadratic programming (QP) optimization~\cite{floudas1995quadratic} is applied to the feature expectations from Eq.~\ref{eq:feat-exp} to find a minimal scale weight parameter such that the constraint is satisfied (Eq. 13 from~\cite{abbeel2004apprenticeship}:
\begin{equation}
\begin{split}
   QP(\mathbf{w}) = min & \quad ||\mathbf{w}||\\
    s.t. & \quad [\mu (\pi_e) - \mu (\pi_l)]^{\intercal} \mathbf{w} \geqslant \epsilon
\end{split}
\end{equation}
where $\mu (\pi_e)$,  $\mu (\pi_l)$ are feature expectations of the expert and learner trajectories respectively, and $\epsilon$ is a pre-defined hyperparameter. After getting an initialized reward function from the QP optimization process, the Q-function is learned through the general temporal difference learning \cite{sutton2018reinforcement} process.

\textbf{MaxEnt-IRL} The second baseline uses another foundational IRL method introduced in \cite{maxentIRL2008}, that uses maximum entropy based approach for deducing the reward function from expert trajectories. The difference with our approach is that we apply the algorithm to the observations, $O$ without conditioning on the structural motif, $U \times O$.

\textbf{f-IRL} To test where our approach can learn good policies in high-dimensional continuous control tasks, we use the f-IRL approach~\cite{f-irl2020} as a baseline that learns a reward function and optimal policy simultaneously using \textit{state marginal matching}.

%%%%%%%%%%%%%%%%%%%%%%%%%%%%%%%%%%%%%%%%%%%%%%%%%%%%%%%%%%%%%%%%%%%%%%%%%%%%%%%% 
\subsection{Office Gridworld}\label{sec:experiments-gridworld}
\begin{figure}[!t]
     \centering
     \begin{subfigure}[t]{0.3\textwidth}
         \centering
         \includegraphics[width=\textwidth]{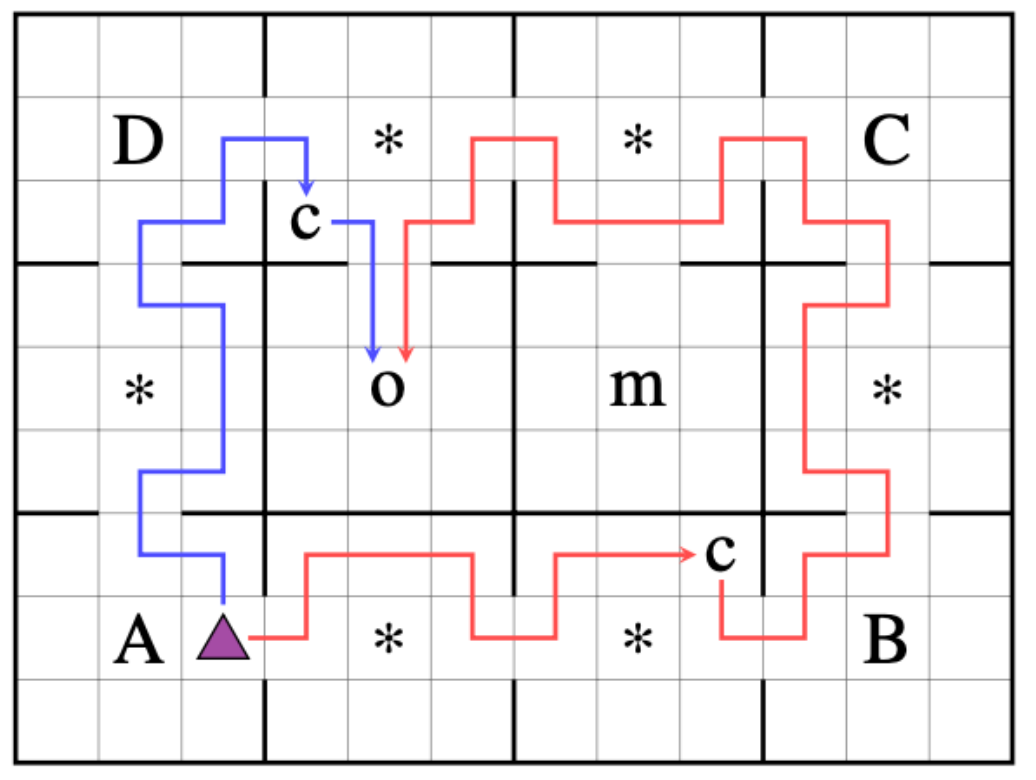}
         \caption{The Office GridWorld }
         \label{fig:office-world-grid}
     \end{subfigure}
     \hfill
     \begin{subfigure}[t]{0.35\textwidth}
         \centering
         \includegraphics[width=\textwidth, height=1.5in]{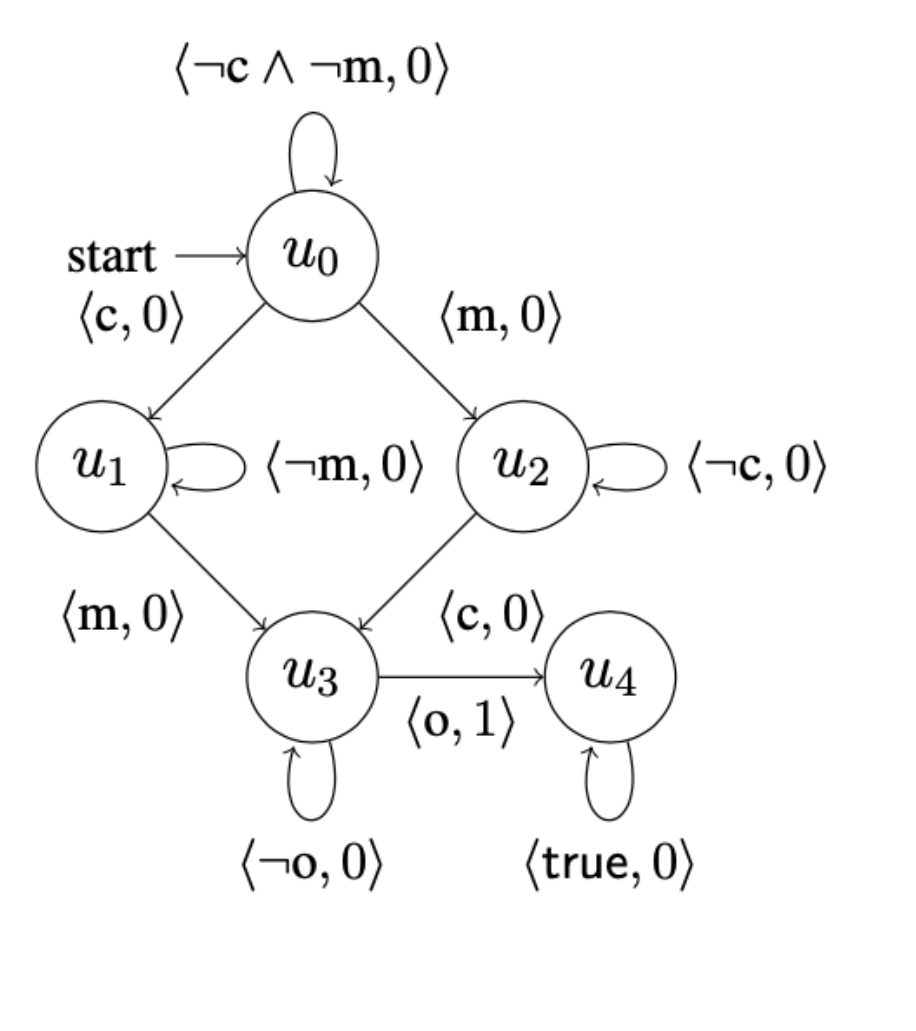}
         \caption{Tasks 1-3: Delivering Coffee \& Mail}
         \label{fig:rm-task13}
     \end{subfigure}
     \hfill
     \begin{subfigure}[t]{0.33\textwidth}
         \centering
         \includegraphics[width=\textwidth]{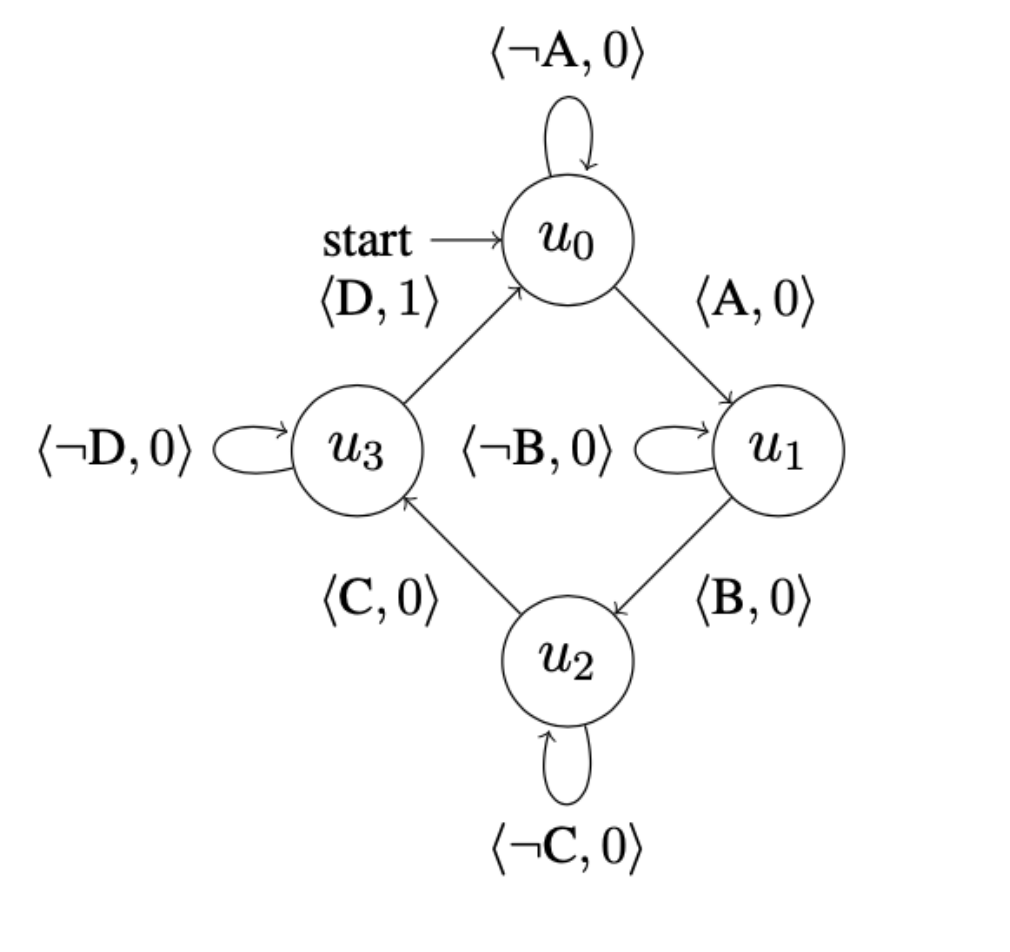}
         \caption{Task 4: Patrol A,B,C, and D}
         \label{fig:rm-task4}
     \end{subfigure}
    \caption{\textbf{The Office GridWorld environment} (images from~\cite{qrm2018}) . In~\textbf{(\ref{fig:office-world-grid})}: the small pink triangle shows the agent's position. The small letters \texttt{o,c,m} shows the location of the office, coffee, and mail respectively. The capital letters \texttt{A,B,C,D} represents the four locations that can be visited by the agent. The asterik \texttt{*} represents an obstacle. The blue and red lines represent the optimal and sub-optimal paths respectively for completing Task 1. Fig.~\textbf{(\ref{fig:rm-task13})}, \textbf{(\ref{fig:rm-task4})} show ground-truth FSA structure for sub-captioned tasks in the Office GridWorld environment. } 
    \label{fig:office-world}
\end{figure}

% We conduct experiments first in a simpler discrete domain, and then in 3D continuous domain. 
% \paragraph{Office Gridworld} 
We use the discrete state `\textit{Office Gridworld}' environment used in~\cite{qrm2018} for our initial experiments. We designed five tasks (referred to as `\textit{Task n}' where $n \in [0-4]$ ) on this domain. Fig.~\ref{fig:office-world} shows the schematic diagram and examples of task specific finite state automatons (i.e. RMs) in this domain.

\textbf{Tasks} 
The first task (\textit{Task 0}) is a simple MDP task designed to verify the efficacy of our proposed algorithm and the foundational IRL baselines. With a single destination location \texttt{D} on the map, the goal for the agent is to reach \texttt{D} without going over any obstacle. The other four tasks are sequential non-Markovian tasks used in \cite{qrm2018}. These tasks can be dichotomized into \textit{delivery} and \textit{patrol} tasks. In \textit{Task 1} (resp. \textit{Task 2}), the agent is expected to fetch and deliver coffee (resp. the mail) to the office. In \textit{Task 3}, the agent is expected to deliver both. \textit{Task 4} requires the agent to patrol or visit the grid in a specified sequential route: $A \xrightarrow{} B \xrightarrow{} C \xrightarrow{} D $ in order to receive a reward. Figures~\ref{fig:rm-task13},~\ref{fig:rm-task4} show the perfect RM structures that describes tasks 3 \& 4.

\textbf{Generating Expert Trajectories} 
We generate test-time demonstration trajectories based on the implementation \footnote{\url{https://bitbucket.org/RToroIcarte/qrm/src/master/}} attached to the original QRM paper \cite{qrm2018}. We make sure that the saved demonstrations are optimal ones by counting the number of steps before the trajectories end. 

The expert demonstrations are represented by series of observation-action transitions, i.e., $(o_1, a_1, o_2, a_2, ..., o_T)$. The actions are represented by single integers from $0$ to $3$. The observations contain both the raw observation from the environment (the agent's position on the grid) as well as the output of the labelling function. In the \textit{Office Gridworld}, the labelling function $\mathcal{L}$ outputs the label of the tile the agent is on (i.e., `A' if the agent is on position \texttt{A}). We refer to this \textit{labels} as `\textit{high-level features}' in subsequent parts.

\begin{figure}[!t]
\begin{minipage}[c]{0.50\textwidth}
  \centering
  \includegraphics[width=\textwidth]{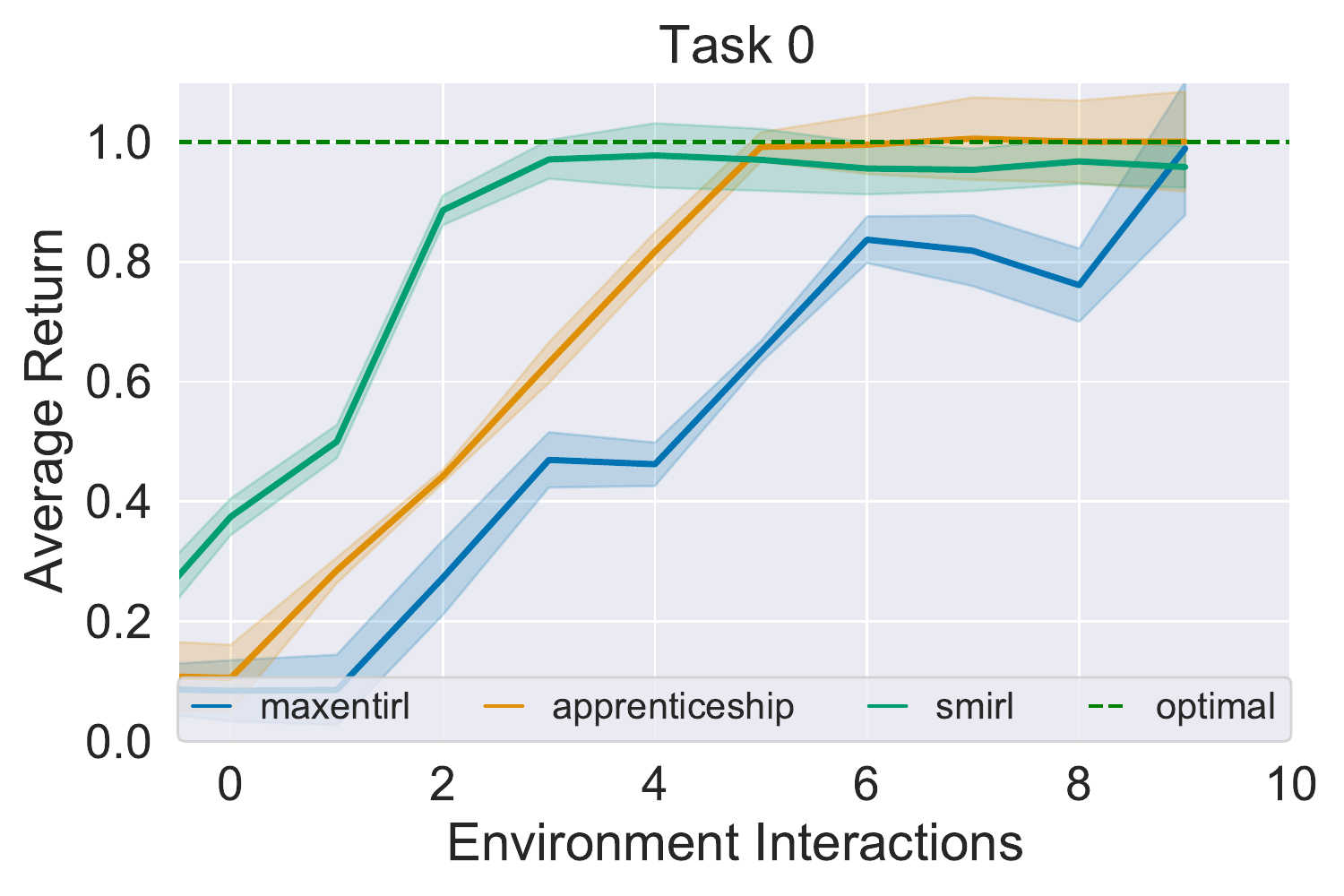}
\end{minipage}
% \,
\begin{minipage}[c]{0.48\textwidth}
\centering
\caption{\textbf{Single-Step MDP (Task 0)}: reach location \texttt{D} avoiding obstacles. All algorithms succeed in this pedagogical example. A final reward of 1 shows successful task objective completion. Ours (in green) is sample efficient and always takes the optimal (\textcolor{blue}{blue}) path in Fig.~\ref{fig:office-world-grid} due to the underlying causal structural motif. }
\label{fig:results_t0}
\end{minipage}
\end{figure}

% %%%%%%%%%%%%%%%%%%%%%%%%%%%%%%%%%%%%%%%%%%%%%%%%%%%%%%%%%%%%%%%%%%%%%%%%%%%%%%%%
\textbf{Performance in Single Step Tasks}
Fig.~\ref{fig:results_t0} shows the results of the simple MDP task (\textit{Task 0}) -- where the agent simply needs to reach the location \texttt{D} from its starting position while avoiding the obstacles. The average reward that the learned policy gets from the ground truth reward function is shown at each IRL step (i.e., each time a new reward function is learned). As we can see, all three algorithms are able to learn an optimal policy on this task, although our algorithm learns in fewer steps. 
% A final reward of 1 here shows that the agent is able to successfully reach the destination. 

This faster learning can be explained by the terminal (goal) state in the reward machine. We prevent the environment from resetting to obfuscate the real reward function from the agent. Thereby, during training, the environment does not reset upon task completion (reaching \texttt{D}) or `game over' (going over an obstacle). Thus, our approach learns the optimal or maximum reward state by inferring the goal state (from the learned FSA), while the other algorithms don't. Also, having the causal structural motif always ensure ours to take the optimal path to success unlike the baselines.
% The drop in performance for MaxEnt at IRL step 6 is due to the agent going over \texttt{A} at first when going to \texttt{D}. It causes a negative reward for \texttt{A} which then through the normalization of the reward function lowers the penalty for going over an obstacle, which thus becomes lower than the shortfall from going around it.

\begin{figure}[!t]
     \centering
     \begin{subfigure}[t]{0.48\textwidth}
         \centering
         \includegraphics[width=\linewidth]{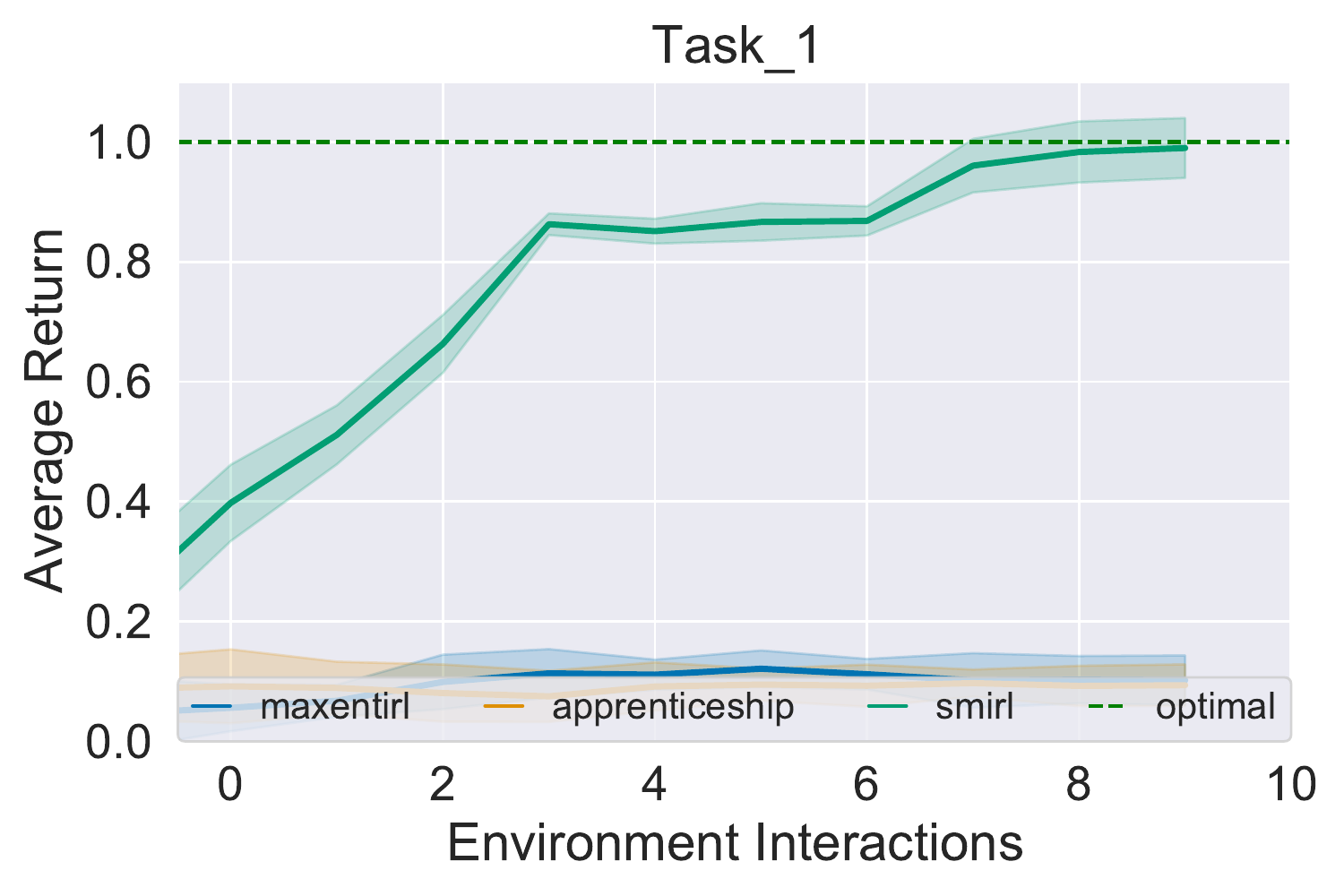}
         \caption{Task 1: Fetch \& Deliver `Coffee'} \label{fig:task1}
     \end{subfigure}
     \hfill
     \begin{subfigure}[t]{0.48\textwidth}
         \centering
         \includegraphics[width=\linewidth]{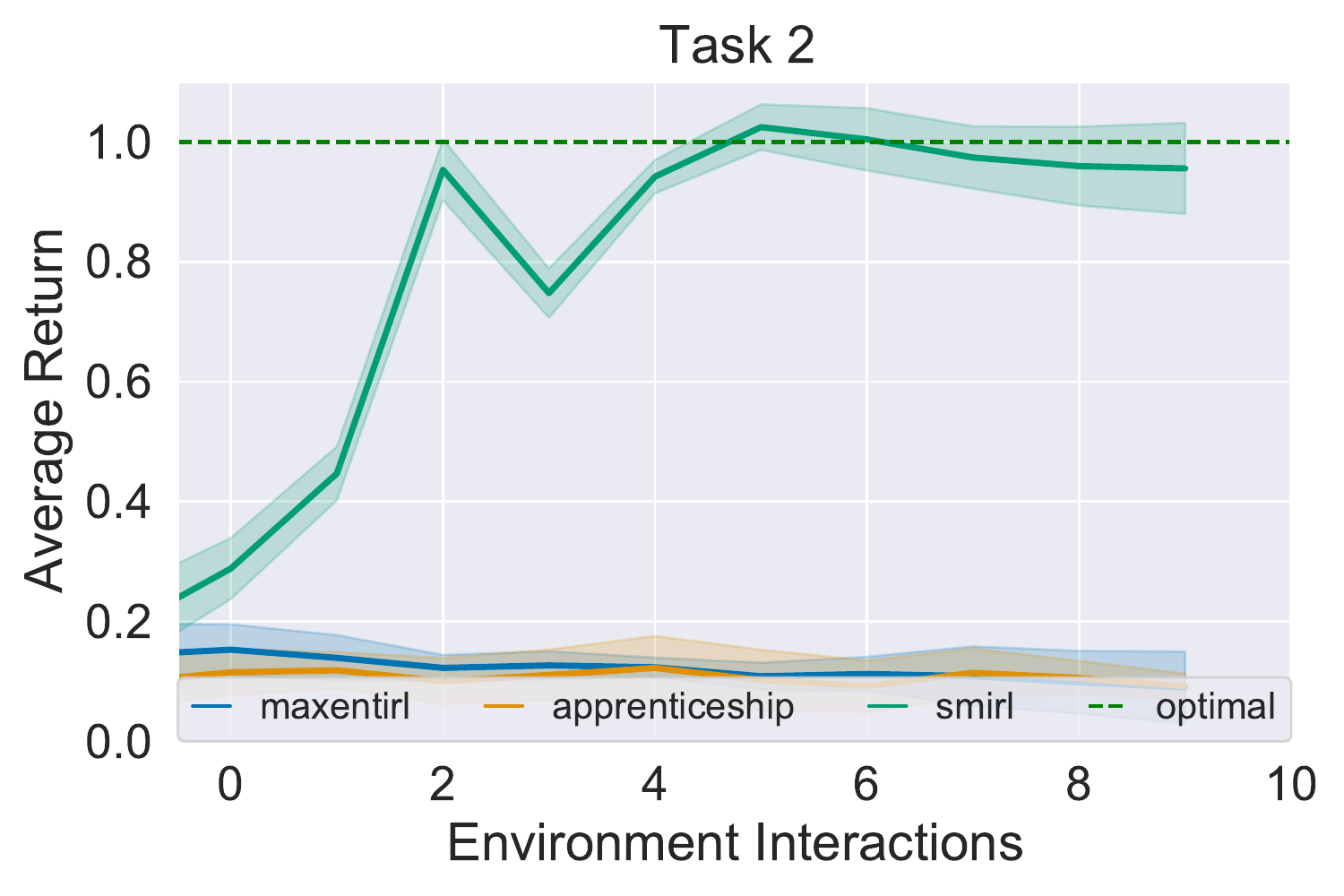}
         \caption{Task 2: Fetch \& Deliver `Mail'} \label{fig:task2}
     \end{subfigure}
     \hfill 
     \begin{subfigure}[t]{0.48\textwidth}
         \centering
         \includegraphics[width=\textwidth]{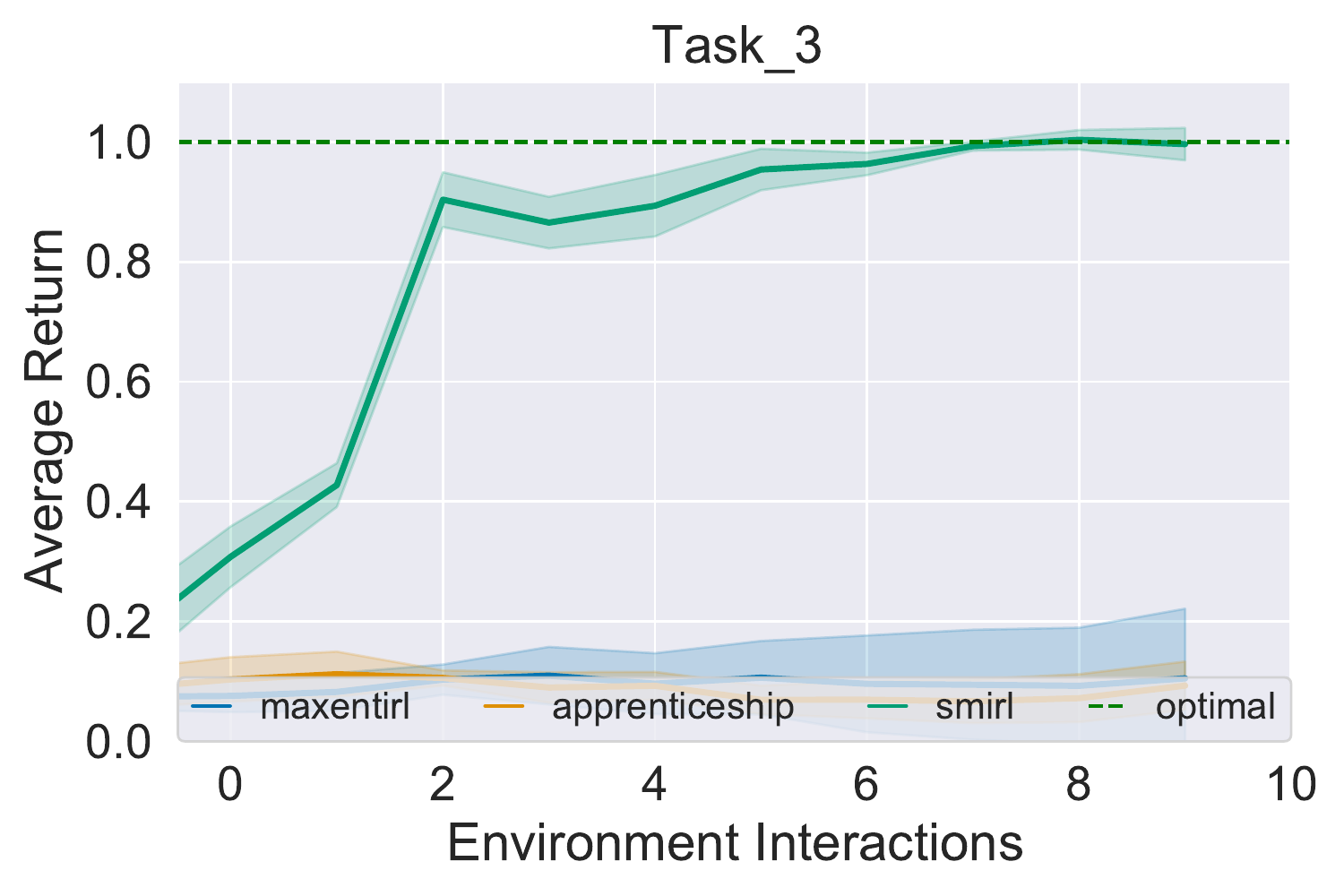}
         \caption{Task 3: : Fetch \& Deliver `Coffee' and `Mail'} \label{fig:task3}
         \label{fig:rm-IF}
     \end{subfigure}
     \hfill
     \begin{subfigure}[t]{0.48\textwidth}
         \centering
         \includegraphics[width=\textwidth]{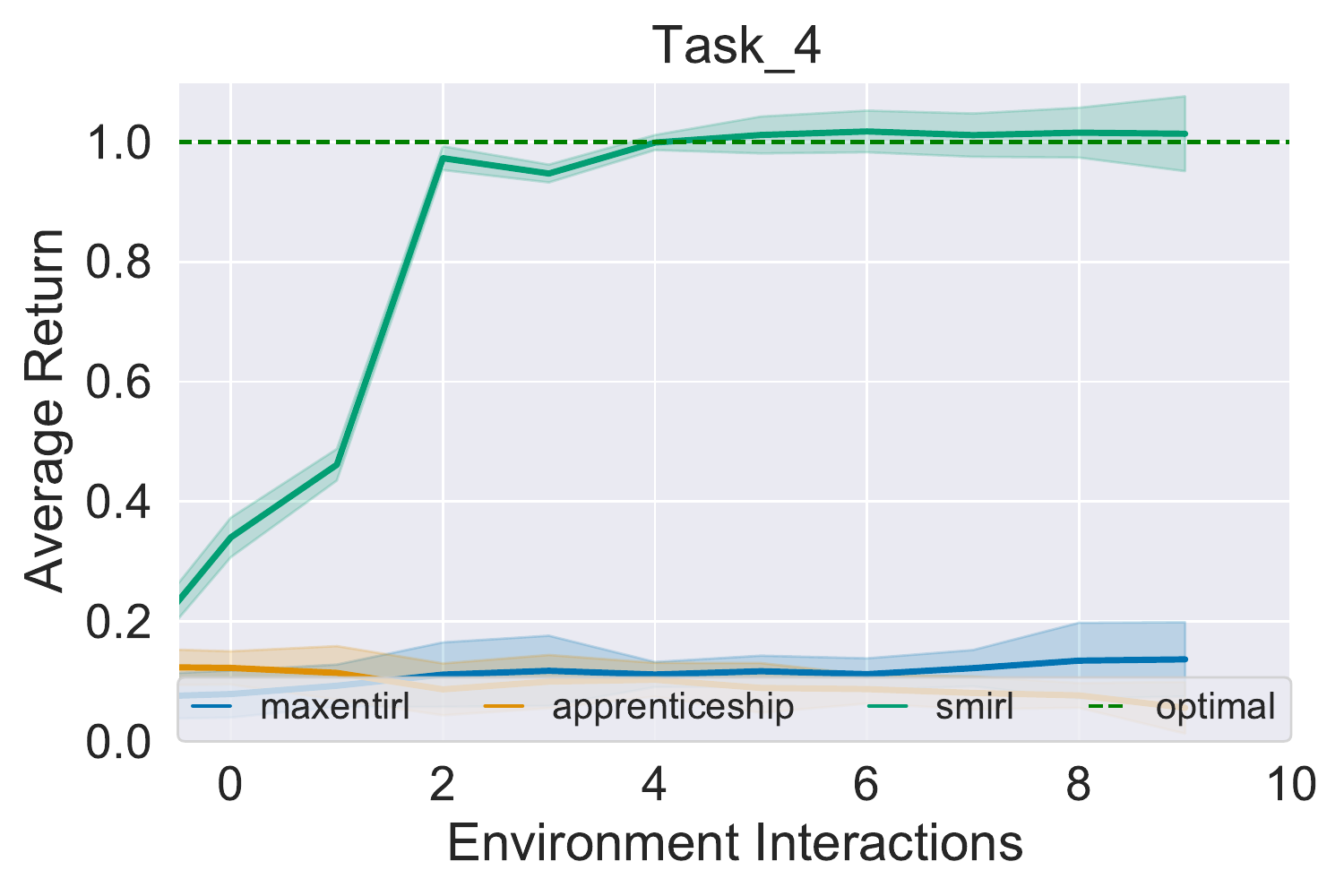}
         \caption{Task 4: Four-Point Sequential Patrol} \label{fig:task4}
     \end{subfigure}
    \caption{\textbf{Multi-step Partially-Observable Domains} training curves for SMIRL and baselines on the POMDP tasks in \textbf{Office GridWorld} environment. All Environment interactions (x-axis) are $1 \times 10^5$. Reward of 1 is obtained when target location `o' is reached after satisfying conditional goals without environmental rewards. SMIRL learns the underlying structure with state-transition and reward functions $\delta_u, \delta_r$. }
    \label{fig:gridworld-results}
\end{figure}

% \animesh{this is the key winning result, provide more technical intuition.}
\textbf{Performance in Multi-step Partially-Observable Domains (1-4)} 
However, in the more complicated sequential tasks shown in Fig.~\ref{fig:gridworld-results}, only SMIRL succeeds to learn the tasks while the other IRL algorithms fail. This shows that our algorithm is able to learn in a POMDP setting, by learning an adequate causal structure (i.e. a RM) from the expert trajectories. An example learned RM can be seen in Fig.~\ref{fig:app-gridworld-learned-rm}. In these tasks, an reward of 1 is given only when a conditional logic has been met prior to target goal. To elucidate, the delivery tasks (Tasks 1-3) require the agent reaching coffee or mail or both locations prior to reaching the target location: office `o'. Thus, from the $\mathcal{D}_e$ trajectories, policies observe the reward after reaching `o', the dependency on the prior conditions make these tasks complex of POMDP nature. Here, ours learn the underlying structure, and corresponding state-transition reward function $\delta_r : U \to [S \to \mathbb{R}]$ that allows succeeding at these tasks.
% \textit{Project code is available at }~\footnote{\url{https://github.com/realrabot/irl-motifs}}

%% Moved Fig 5. to the appendix.
% \begin{figure}[!t]
% \begin{minipage}[c]{0.68\textwidth}
%   \centering
%   \includegraphics[width=\textwidth]{imgs/IRL/reward_function_t3.pdf}
% \end{minipage}
% % \,
% \begin{minipage}[c]{0.3\textwidth}
% \centering
% \caption{\textbf{Qualitative Evaluation in Office World}. The learned reward machine and weights learned for Office Gridworld \textit{Task 3: Fetch \& Deliver `Coffee' and `Mail'}. We can see some artifacts of the LRM algorithm here with nodes $u_2$ and $u_4$. The first transition from $u_0$ to $u_1$ comes from the fact that the expert trajectories go through \texttt{D} while going to \texttt{c}.}
%     \label{fig:reward_function_t3}
% \end{minipage}
% \end{figure}

\subsection{Reacher Domain}\label{sec:experiments-reacher}
\begin{wrapfigure}[17]{R}{0.45\textwidth} 
    \centering
    \hspace{6pt}
    % \vspace{-6pt}
    %\vspace{-\baselineskip}
    \includegraphics[width=0.45\textwidth]{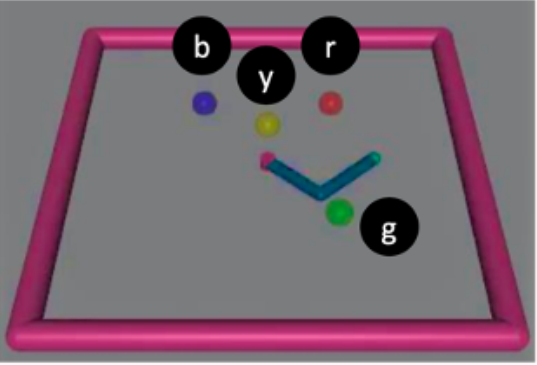}
    \vfill \vspace{3pt}
    \caption{ReacherDelivery-v0: the colored balls specify the various target locations, with the black circles depicting corresponding labels. The arm's \textit{fingertip} represents the agent's location.}\label{fig:reacher-domain}
    % \vspace{-6pt}
\end{wrapfigure}

We use a modified version of the reacher domain (Fig.~\ref{fig:reacher-domain}) from OpenAI Gym~\cite{brockman2016openai}. It is a two-jointed (2 DOF) robot arm, and the goal is to move the arm's end effector (or `\textit{fingertip}') close to the target location(s). The original environment spawn a target at a random position. For our purposes, we use a modified version with four additional target positions denoted by colored balls to induce subgoals for complex trajectories.

\textbf{Tasks} Four tasks were designed to satisfy four logical conditions of increasing difficulty. \textit{Task OR} requires the agent to touch "\textit{the red or green ball, and then the blue ball}". \textit{Task IF} stipulates to touch "\textit{the blue ball, unless a cancel event occurs, then go to red}". The modified environment generates a random `cancel' event to facilitate this task. \textit{Task Sequential} is equivalent to a patrol task which requires the agent to touch "\textit{the red, blue, green, and yellow balls in order}". Lastly, \textit{Task Composite} is a hybrid task that composes all the preceding logics by requiring the agent to "\textit{go to red or green, then blue unless cancel event, then go to yellow}".  
% (State diagrams will be submitted with supplementary materiasl. TODO. Insert perfect FSA tasks' structures as (b) above

\textbf{Generating Expert Trajectories} Expert trajectories were generated by training a f-irl~\cite{f-irl2020} model with forward KL divergence (fkl) with perfect reward structure using SAC~\cite{sac2018}. Implementation details of the expert training and trajectories are detailed in Appendix~\ref{app:expert-trajectories}. 16 expert trajectories generated by the fully trained expert were used for carrying out experiments in this section.

\begin{figure}[!t]
     \centering
     \begin{subfigure}[t]{0.48\textwidth}
         \centering
         \includegraphics[width=\linewidth]{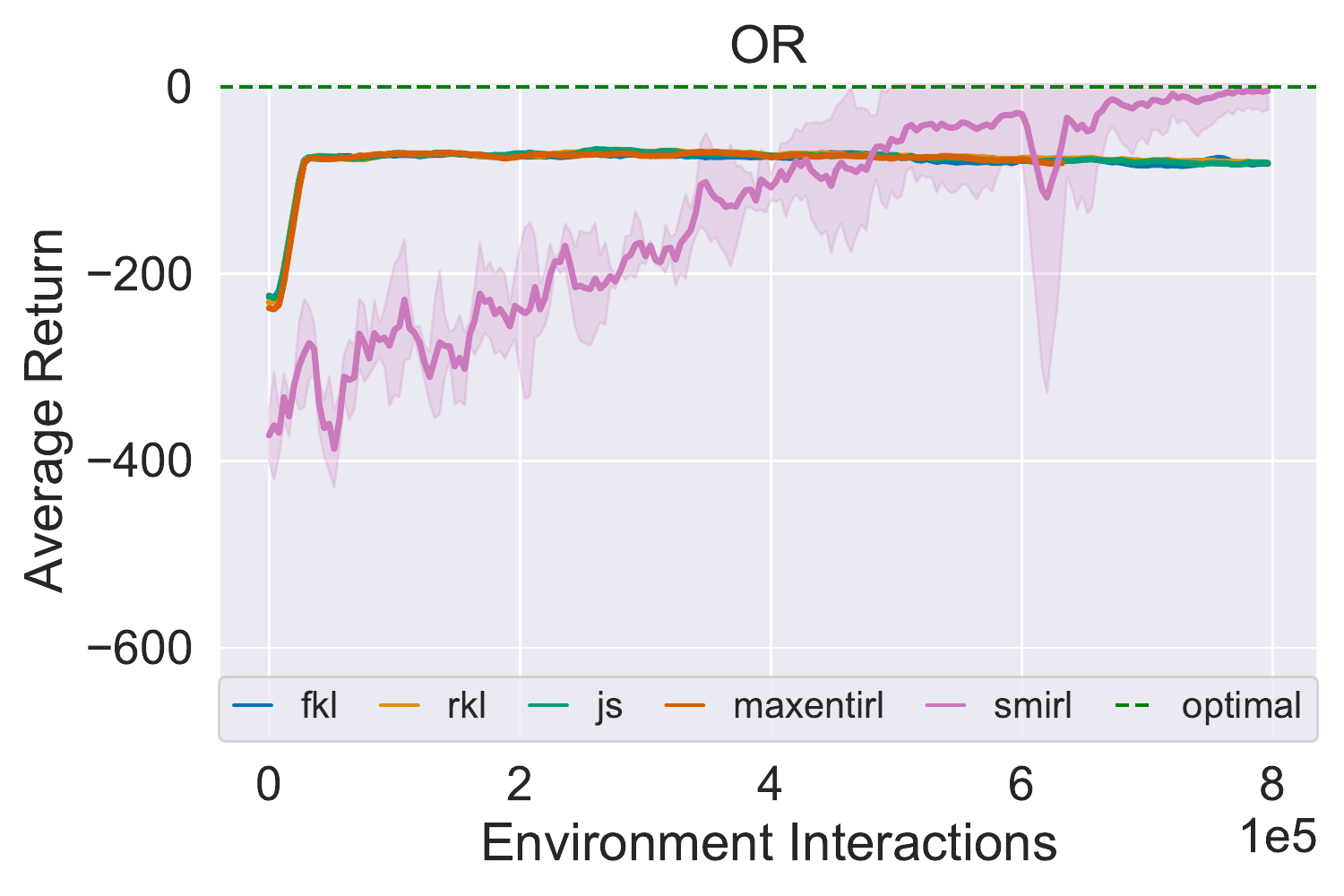}
        %  \caption{Task "OR"} \label{fig:rm-OR}
     \end{subfigure}
     \hfill
     \begin{subfigure}[t]{0.48\textwidth}
         \centering
         \includegraphics[width=\linewidth]{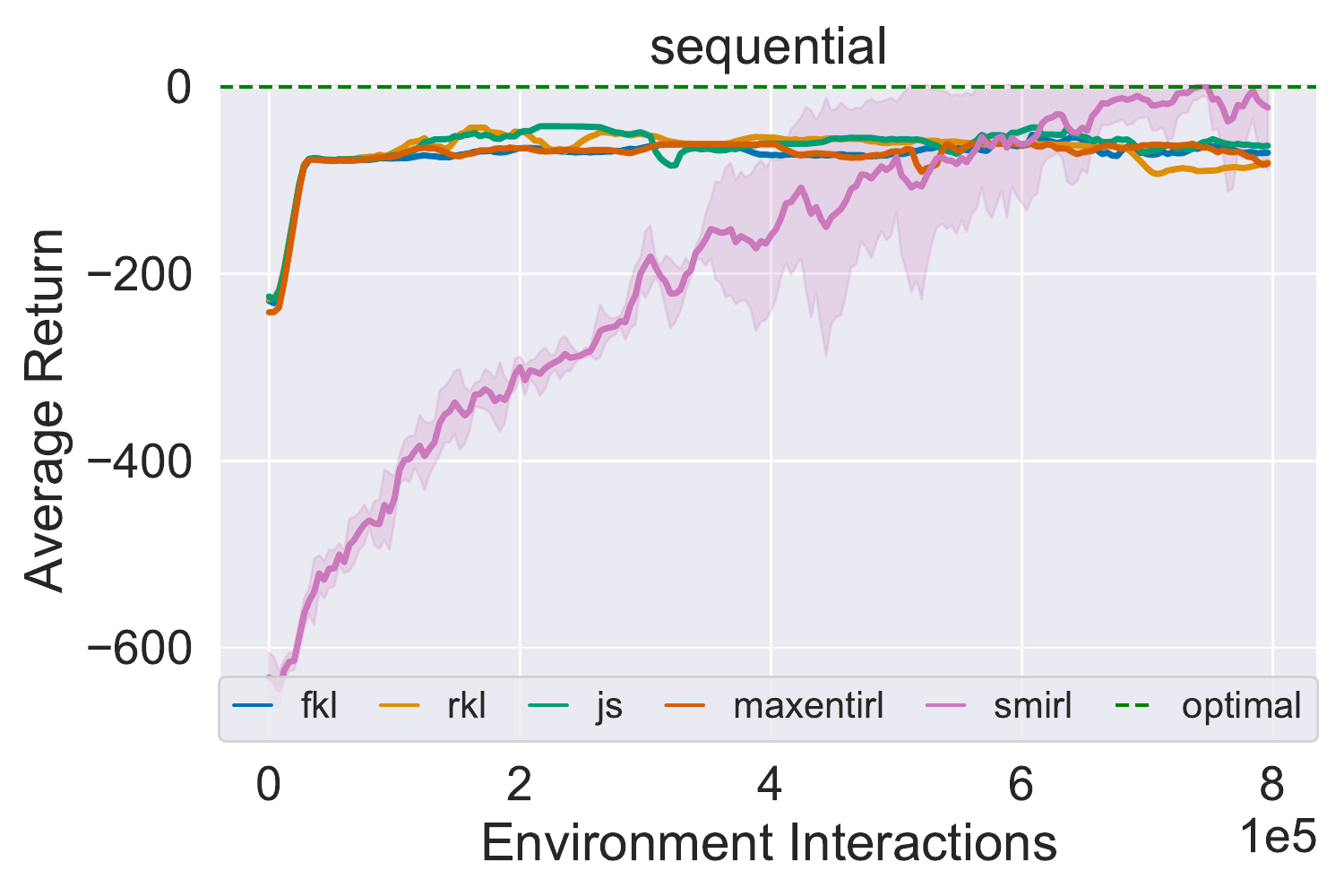}
        %  \caption{Task "sequential" } \label{fig:reacher-seq}
     \end{subfigure}
     \hfill 
     \begin{subfigure}[t]{0.48\textwidth}
         \centering
         \includegraphics[width=\textwidth]{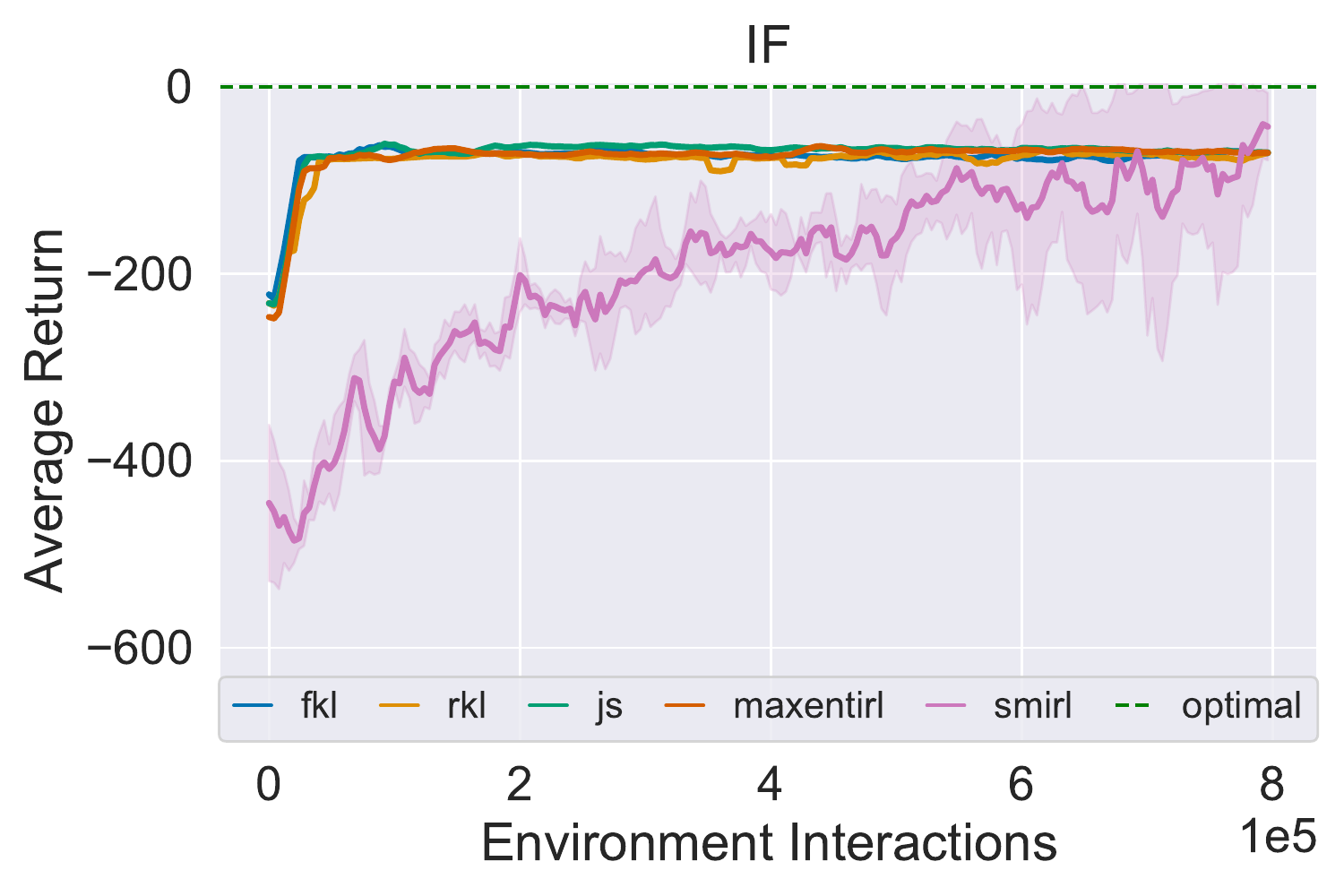}
        %  \caption{Task "IF"}\label{fig:rm-IF}
     \end{subfigure}
     \hfill
     \begin{subfigure}[t]{0.48\textwidth}
         \centering
         \includegraphics[width=\textwidth]{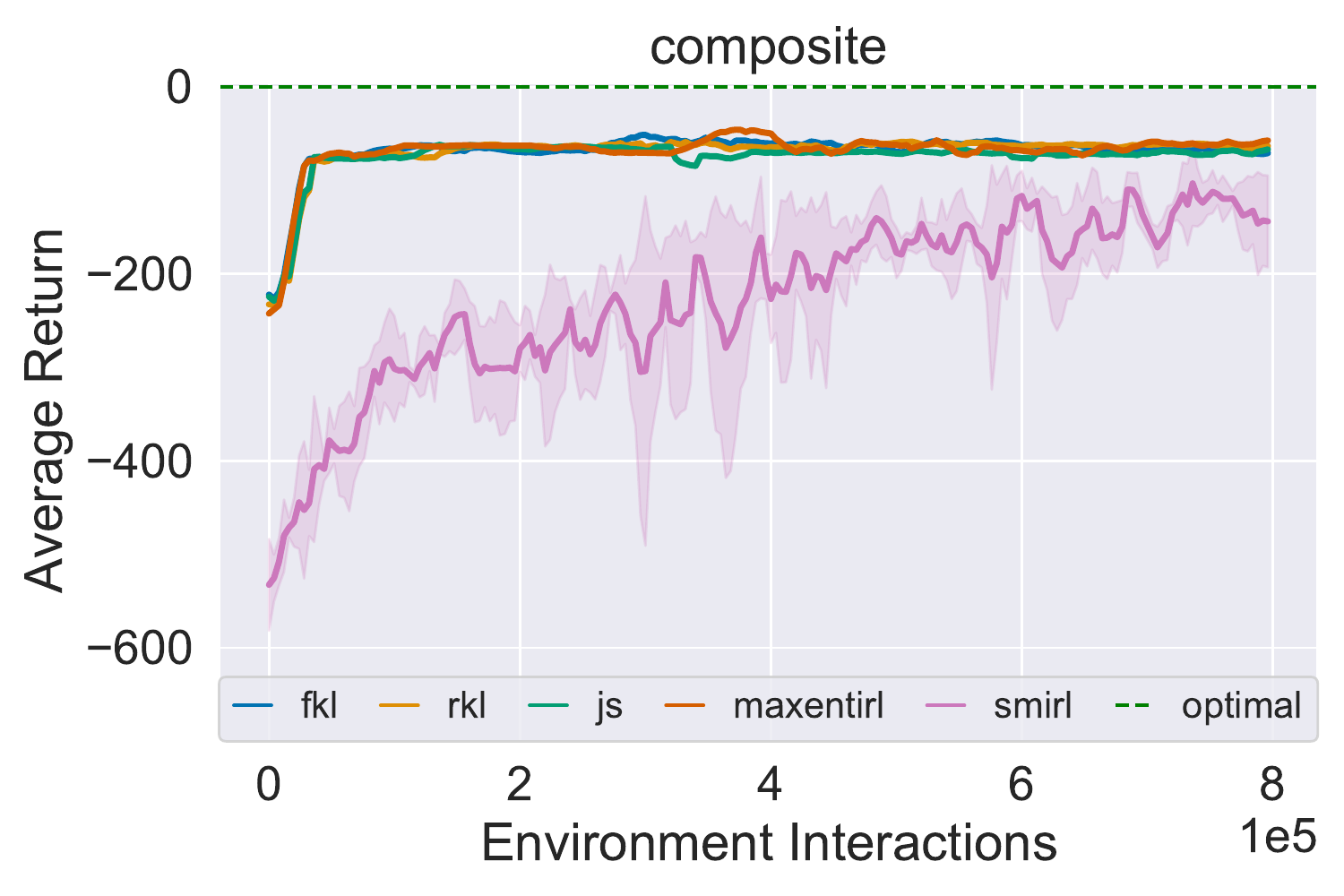}
        %  \caption{Task "COMPOSITE" } \label{fig:rm-COMPOSITE}
     \end{subfigure}
    \caption{Training curves for our system (\texttt{SMIRL}) and baselines on the \textbf{ReacherDelivery} environment. Solid curves and shaded areas depict the mean and standard deviation of three trials with different random seeds respectively for each of the four tasks.}
    \label{fig:reacher-results}
\end{figure}

\textbf{Results} Fig.~\ref{fig:reacher-results} shows the training curves of our system and baselines. The baselines saturate at a partial reward level, given for reaching the end location; however, learning the optimal reward requires learning to satisfy prior conditions (specific order as stipulated by the task's logical constraints), which none of the baselines are able to recover. With our approach, the agent successfully learns the optimal reward function in all 3 tasks barring the last and most difficult composite task. The perfect FSA structure for Task 4 is 7 states with complex logical constraints among them. We suspect that our model converges on an erroneous sub-optimal reward structure while learning a RM with large state space. Thus, it underperforms against sub-optimal baseline agents in the composite task. 

%%% HORIZONTAL STYLE (1,4) %%

\section{Limitations \& Future Work} \label{sec:lims}
Comprehensive exploration of IRL in POMDPs is beyond the scope of our work. In general, it is intrinsically an ill-posed problem (due to the nature of IRL) and also can be computationally intractable due to the difficulty in solving POMDPs. Choi \& Kim~\cite{choi2009} included a detailed exploration in this setting, and future work might compare our model's performance against the three approaches that use sampled trajectories (Max-margin between values (MMV), Max-Margin between feature expectations (MMFE), and the projection (PRJ) methods) and corresponding domain experiments. However, we notice that these methods are derivatives or extensions of algorithms outlined in \cite{abbeel2004apprenticeship, ng2000algorithms}, therefore they may have been subsumed by our baselines used, making a proxy comparison with these approaches. 

A limitation of our model is the intrinsic limitation or weakness of using RM (or FSA-based abstractions) as structural motifs. While we don't handcraft domain specific structures (they are learnt), but the learning process itself is dependent on a domain-specific labelling function $\mathcal{L}$ assumption. Such perfect domain sensing may not be generalize to all environments.

% Essentially, we omit the case of noisy observations and assume perfect sensing which, while limiting, is plausibly a fair hard assumption to make for modelling purposes. 

% One criticism of using RM is the need for handcrafting domain specific reward structures. In our approach, we circumvent this issue by learning the structure,
% Future work should similar explore comparisons with generative probabilistic models \cite{hahn2015} and generative adversarial methods \cite{henderson2018}.

% An obvious limitation is the dearth of adequate experiment domains considered. Given the limited time-frame of the project, we only considered a simpler discrete world domain for our experiments. We plan to extend and conduct more experiments on additional discrete and continuous domain environments (like OpenAI Gym~\cite{brockman2016openai}) to verify the efficacy of our proposed model.
% Fairer comparison w/ IRL-POMPD methods.
% The second set of experiments involves the case when the expert’s trajectories are given. We exper- imented on the same set of five problems with all three approaches in Section 6: the max-margin between values (MMV), the max-margin between feature expectations (MMFE), and the projection (PRJ) methods.
\newpage
\section{Broader Impact and Conclusion}\label{sec:conclusion}
In this work, we proposed a novel approach to learn an agent's reward function from expert observations (i.e., the IRL problem) by inducing structural motifs (in the form of learned reward machines). We empirically show that our approach (SMIRL) is able to successfully learn complex (non-Markovian) sequential tasks in both discrete grid world and high-dimensional continuous domain  environments, where foundational and SOTA IRL methods fail. These results are highly motivating for further applications in continuous domain like robotics. Especially, our approach is highly promising for the RoboNLP domain, with the vast swath of rich literature on parsing logical structures from natural language, thus enabling the inverse mapping of agent (robot) behavior to natural language descriptions.  

% The results obtained in the simple (discrete) domain encourages us to conduct further experiments and analysis on more complex (continuous) domain tasks to conclusively demonstrate the efficacy of our proposed novel approach. 

\newpage
\bibliographystyle{plain} %  {plainnat}
\bibliography{main, bibs/irl, bibs/rebuttal}

%%%%%%%%%%%%%%%%%%%%% APPENDIX %%%%%%%%%%%%%%%%%%%%%%%%%
\newpage
% \subfilestrue
\ifsubfiles
    \documentclass[./00-main.tex]{subfiles}
    \usepackage{graphicx}
    \usepackage{array}
    \begin{document}
    
\fi
%%%%%%%%%%%%%%%%%%%%%%%%%%%%%%%%%%%%%%%%%%%%%%%%%%%%%%%%%%%%%%
%%%%%%%%%%%%%%%%%%%%% APPENDIX START %%%%%%%%%%%%%%%%%%%%%%%%%
%%%%%%%%%%%%%%%%%%%%%%%%%%%%%%%%%%%%%%%%%%%%%%%%%%%%%%%%%%%%%%
\appendixpage
\appendix
%%%%%%%%%%%%%%%%%%%%%%%%%%%%%%%%%%%%%%%%%%%%%%%%%%%%%%%%%%%%%%%%%%%%%%
%%%%% APPENDIX A:  %%%%%%%%%%%%%%%%%%%%%%%%%%%
%%%%%%%%%%%%%%%%%%%%%%%%%%%%%%%%%%%%%%%%%%%%%%%%%%%%%%%%%%%%%%%%%%%%%%
\section{Definitions}\label{app:supp-materials}
\begin{defn}[Domain vocabulary \& labeling function]{def:domain.vocab}
A \textit{vocabulary} is a set of environment or domain specific (boolean) propositional symbols, $\mathcal{P}$, with $\mathbb{R} \in [0,1]$ for each state $s \in \mathcal{S}$. 
A \textit{labeling function} is a function $\mathcal{F}_L(t): \mathcal{S} \times \mathcal{A} \times \mathcal{S'} \rightarrow 2^\mathcal{P}$, that maps state observations to truth assignments over the vocabulary $\mathcal{P}$ for the time-step $t$.
\end{defn}

\begin{defn}[MDP with a reward machine (MDP-RM)]{def:MDPRM}
A Markov decision process with a reward machine is a tuple $\Psi = \langle S, A, p, \gamma, \mathcal{P}, \mathcal{F}_L, U, u_0, G, \delta_u, \delta_r \rangle$ where $S, A, p$, and $\gamma$ are defined as in an MDP, $\mathcal{P}$ is a set of domain specific propositional symbols, $\mathcal{F}_L$ is a labelling function (Def.~\ref{def:domain.vocab}), and $U, u_0, G, \delta_u$ and $\delta_r$ are defined as in a reward machine, where U is a set of FSA states, $u_0$ is the initial state, $G$ is a set of terminal or goal states, and $\delta_u, \delta_r$ are state and reward transition functions.
\end{defn}

% \todo{add proof of convergence for \ref{alg:LRM-MaxEntIRL}} solving M_rm should solve the problem.

%%%%%%%%%%%%%%%%%%%%%%%%%%%%%%%%%%%%%%%%%%%%%%%%%%%%%%%%%%%%%%%%%%%%%%
\section{Reward Machine: Expanded Background and Motivation }\label{app:reward-machine}
%%%%%%%%%%%%%%%%%%%%%%%%%%%%%%%%%%%%%%%%%%%%%%%%%%%%%%%%%%%%%%%%%%%%%%
\textbf{Reward machines} were introduced by Icarte et al.~\cite{qrm2018} as a type of finite state machine (FSM) that supports the specification of reward functions while exposing reward function structure. As a form of FSM, reward machines have the expressive power of a regular language and as such, support loops, sequences and conditionals. Additionally, it supports expression of temporally extended linear-temporal-logic (LTL) and non-Markovian reward specification, where the underlying reward received by an agent from the environment is not Markovian with respect to the state.

When an environment dynamics is specified using a reward machine (RM), as an agent acts in the environment, moving from state to state, it also moves from state to state within a reward machine (as determined by high-level events detected within the environment).

After every transition, the reward machine outputs the reward function the agent should use at that time. For example, we might construct a reward machine for `\textit{delivering mail to an office}' using two states. In the first state, the agent does not receive any rewards, but it moves to the second state whenever it gets the mail. In the second state, the agent gets rewards after delivering the mail. Intuitively, defining rewards this way improves \textit{scale efficiency} as the agent knows that the problem consists of two stages and might use this information to speed up learning.

Using the above discussion, we can define a standard RM using mathematical formalism as the following definition~\ref{def:RM}.

\vspace{5pt}
\begin{defn}[A standard Reward Machine (RM)]{def:RM}
Given a set of propositional symbols $\mathcal{P}$, a set of (environment) states $S$, and a set of actions $A$, a reward machine is a tuple:
$\mathcal{R}_\mathcal{PSA} = \langle U, u_0, F,\delta_u, \delta_r\rangle$ where $U$ is a finite set of states $u_0 \in U$ is an initial state, $F$ is a finite set of terminal states (where $U \cap F = \emptyset $), $\delta_u$ is the state-transition function s.t. $\delta_u: U \times 2^\mathcal{P} \rightarrow U \cup F$, and $\delta_r$ is the state-reward function, s.t. $\delta_r: U \rightarrow [S \times A \times S \rightarrow \mathbb{R}]$.
\end{defn}

% \begin{defn}[A simple Reward Machine (RM)]{def:simpleRM}
% Given a set of propositional symbols $\mathcal{P}$, simple reward machine is a tuple:
% $\mathcal{R}_\mathcal{P} = \langle U, u_0, \mathcal{F}_L,\delta_u, \delta_r\rangle$ where $U, u_0, F$, and $\delta_u$ are defined as in a standard RM (defn.~\ref{def:RM}), but the state-reward function $\delta_r: U \times 2^\mathcal{P} \rightarrow \mathbb{R}$ depends on $2^\mathcal{P}$ and returns a number instead of a function. 
% \end{defn}

\subsection{Motivation for Using RM in IRL+POMDP Setting}
Agents in modern, real-world RL datasets (e.g. robotics, embodied-ai~\cite{shridhar2020alfred}) often, if not always, are required to perform tasks that are long-horizon with compositional and/or logical underlying reward structure. The main motivation of SMIRL is to show that imbuing the agent with \textit{apriori} (approximate) structure of the latent reward associated with a task allows solving complex tasks that are hard to learn from only demonstrative expert trajectory data. This motivation makes the IRL+POMDP problem setting ideal for purposes in this paper.      

The effectiveness of automata-based memory has long been recognized in the POMDP literature~\cite{cassandra1994acting}, where the objective is to find policies given a complete specification of the environment. The overarching idea in approaches under this umbrella is to encode policies using Finite State Controllers (FSCs), which are FSMs with states associated with one primitive action, and the transitions are defined in terms of low-level observations from the environment. During environment interaction, the agent always selects the action associated with the current state in the FSC. Using such automata-based memory was leveraged to work in the RL setting in work by Meuleau et al.~\cite{meuleau2013learning} by exploiting policy gradient to learn policies encoded as FSCs. 

RMs can be considered as a generalization of FSC as they allow for transitions using conditions over high-level events and associate complete policies (instead of just one primitive action) to each state. This allows RMs to leverage existing deep RL methods to learn policies from low-level inputs, such as images, which is not achievable by other automata-based approaches like~\cite{meuleau2013learning}. That said, learning FSMs using other ontologies (e.g.~\cite{xu2020joint, zhang2019learning}) do exist in concurrent literature. Discussion and delineation with such works are further discussed in the `Related Work' section (S.~\ref{app:related-work}).

% \RS{In essence, the motivation of this work is to show that the well known concept of using reward structures in IRLs can be applied in continuous real world domains. Most work show this as promising in toy 2D grid world domains.}
% The preceding discussion elucidates on why using such a reward function structure is advantageous for environments with underlying latent structure.
%%%%%%%%%%%%%%%%%%%%%%%%%%%%%%%%%%%%%%%%%%%%%%%%%%%%%%%%%%%%%%%%%%%%%%
\section{Related Work}\label{app:related-work}
%%%%%%%%%%%%%%%%%%%%%%%%%%%%%%%%%%%%%%%%%%%%%%%%%%%%%%%%%%%%%%%%%%%%%%
Augmenting memory using of Recurrent Neural Networks (RNNs) in combination with policy gradient~\cite{jaderberg2016reinforcement, mnih2016asynchronous, schulman2017proximal}
is a common approach in state-of-the-art (SOTA) approaches in the RL+POMDP domain. Other approaches use external neural-based memories~\cite{oh2016control, khan2017memory, hung2019optimizing}. Model-Based Bayesian RL and extension approaches~\cite{doshi2013bayesian, ghavamzadeh2015bayesian, poupart2008model} under partial observability provide a small binary memory to the agent and a special set of actions to modify it. The motivation and idea behind our work here are largely orthogonal to these aforementioned approaches. 

% \textbf{}Value differentiation with ~\cite{lrm2020}: In \cite{lrm2020}.
The work that is closest to ours is by Icarte et al.~\cite{lrm2020} -- where the authors learn RMs for partially observable RL tasks from trajectories. However, efficacy of the approach is shown in 2D discrete domains, with the authors noting the challenge of showcasing them in 3D continuous domain due to the intractability of state space explosion. While~\cite{lrm2020} motivates our choice to use RM as the chosen structural motif architecture (as opposed to other available FSA ontologies), but to the best our knowledge, learning the motif on continuous 3D domains with complex logic (see SMIRL Algorithm) is not undertaken in prior works. While this difference can be argued as meagre for 2D toy like domains, but is significant for continuous domains, because `Tabu search' for state space is computationally infeasible in such complex domains. Thus, this insight is more applicable in solving IRL in real-world robotic or embodied-ai domains commensurate with the motivation of our work.

% \RS{Discuss the related work cited by the 4 reviewers}
Another related sub-domain of works include literature on \textit{learning logic and automata from demonstrations}. These works by problem definition is slightly different to the IRL problem domain we tackle here. The works in this area (e.g. by Vazquez et al.~\cite{vazquez2018learning,vazquez2021demonstration})  infers Boolean non-Markovian rewards, or logical properties of available traces (aka. demonstrations). This is achieved by learning probabilistic densities of demonstrations over an \textbf{existing, apriori} knowledge pool of candidate specifications. In essence, it is a specifications matching problem, or searching for the most probable specification in a pool of candidate specifications. Our work is \textbf{orthogonal to these} in the aspect that we do not have or define the task labels or any apriori structure of the specification. 

% R-FB5Q
% [1] Learning Reward Machines for Partially Observable Reinforcement Learning ~\cite{lrm2020}
% [2] Djeumou et al. Task-Guided Inverse Reinforcement Learning under Partial Information. ~\cite{djeumou2022task}
% [3] Malik et al. An Efficient, Generalized Bellman Update For Cooperative Inverse Reinforcement Learning. \cite{malik2018efficient}
% \cite{kasenberg2017interpretable}

%%%%%%%%%%%%%%%%%%%%%%%%%%%%%%%%%%%%%%%%%%%%%%%%%%%%%%%%%%%%%%%%%%%%%%
%%%%%%%%%%%%%%%%%%%%%%%%%%%%%%%%%%%%%%%%%%%%%%%%%%%%%%%%%%%%%%%%%%%%%%

%%%%%%%%%%%%%%%%%%%%%%%%%%%%%%%%%%%%%%%%%%%%%%%%%%%%%%%%%%%%%%%%%%%%%%
%%%%% APPENDIX B: %%%%%%%%%%%%%%%%%%%%%%%%%%%
%%%%%%%%%%%%%%%%%%%%%%%%%%%%%%%%%%%%%%%%%%%%%%%%%%%%%%%%%%%%%%%%%%%%%%
\section{Office Gridworld Domain}
\subsection{Detailed Illustration of Structural Motif Learning (SMIRL)}
\begin{figure}[ht]
    \centering
    \begin{subfigure}[t]{0.3\textwidth}
        \centering
        \includegraphics[width=\textwidth]{./imgs/env/office-gridworld-b.pdf}
         \caption{State diagram of a \textbf{perfect} FSA }
         \label{fig:app-gridworld-perfect-rm}
    \end{subfigure}
    \hfill
    \begin{subfigure}[t]{0.68\textwidth}
        \centering
        \includegraphics[width=\textwidth]{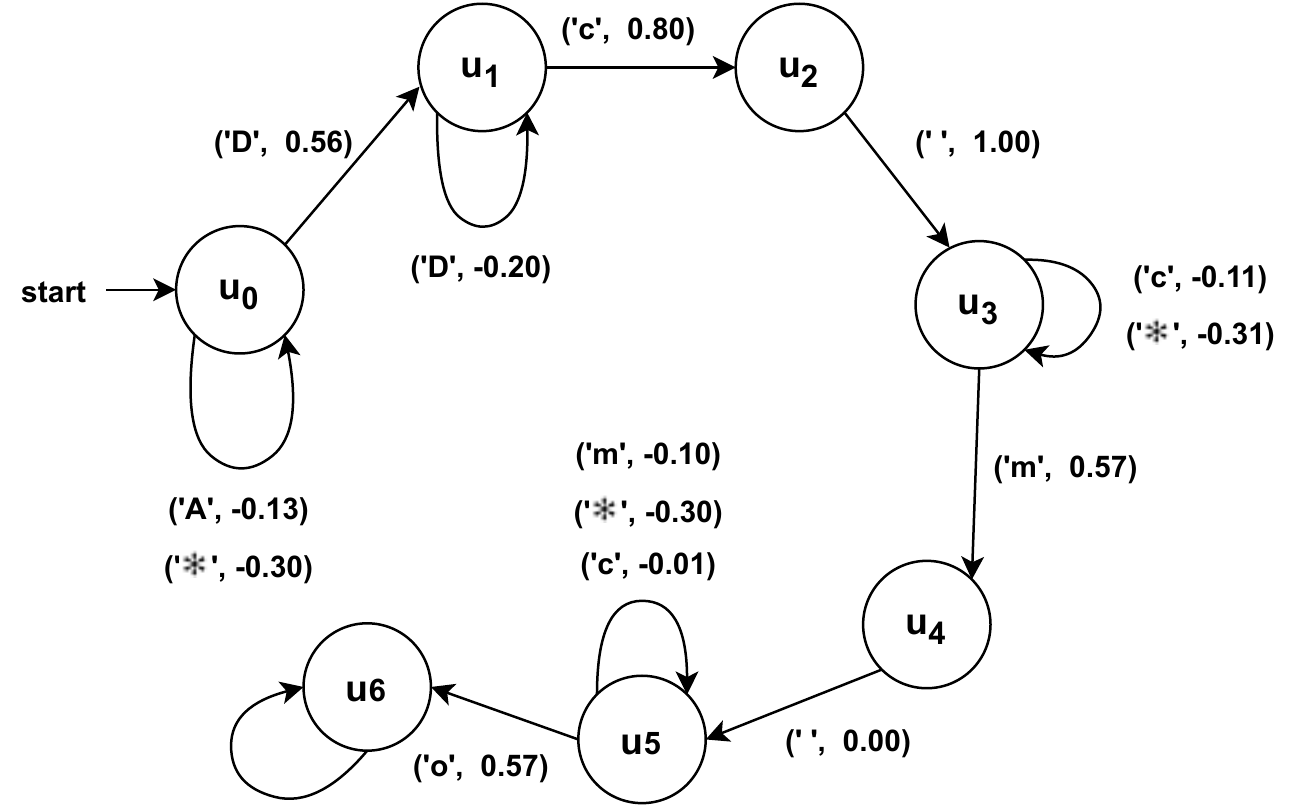}
         \caption{State diagram of \textbf{learned} FSA }
         \label{fig:app-gridworld-learned-rm}
    \end{subfigure}
    \caption{Qualitative Evaluation in the \textit{Office GridWorld}. 
    \textbf{Left}: Fig.~\ref{fig:app-gridworld-perfect-rm} shows the \textbf{perfect} reward (FSA) structure for \textit{Task 3: Fetch \& Deliver `Coffee' and `Mail'} requiring the delivery of both coffee (`\texttt{c}') and mail (`\texttt{m}') to the office (`\texttt{o}') starting with position `\texttt{A}' on the map and initial FSA state $u_0$. 
    \textbf{Right}: Fig.~\ref{fig:app-gridworld-learned-rm} shows the \textbf{learned} FSA structural motif and weights. The state transition arrow labels (e.g. $u_1 \rightarrow u_2$: \texttt{(`c', 0.80)})indicate the true propositional symbols [Def.~\ref{def:domain.vocab}] -- from an agent action $a$ resulting in state transition from $s$ to $s'$, and the underlying FSA to state $u' = \delta_u(u, \mathcal{F}_L(s, a, s')$ -- and the correponding reward $r(s, a, s'$, where $r = \delta_r(u)$.
    We can see some artifacts of the SMIRL algorithm~\ref{alg:LRM-MaxEntIRL} here with nodes $u_2$ and $u_4$. The first transition from $u_0$ to $u_1$ comes from the fact that the expert trajectories go through \texttt{D} while fetching coffee \texttt{c}.}
    \label{fig:app-gridworld-fsa}
\end{figure}

Here we examine and illustrate the SMIRL learning algorithm~\ref{alg:LRM-MaxEntIRL} using \textit{Task 3: `Fetch and deliver coffee and mail`}. The Fig.~\ref{fig:app-gridworld-fsa} juxtaposes the perfect reward structure with a FSA structural motif learned using the SMIRL~\ref{alg:LRM-MaxEntIRL} algorithm. Although the learned structure is not optimal (with 6 FSA states as opposed to optimal 4), but if converges to the desired target state.

\newpage
\begin{lem}[MDP, MDP-RM expected reward equivalency]{lem:MDPRM}
Given an MDP-RM $\Psi = \langle S, A, p, \gamma, \mathcal{P}, \mathcal{F}_L, U, u_0, G, \delta_u, \delta_r \rangle$
let $\mathcal{M}_{\Psi}=\left\langle S^{\prime}, A^{\prime}, r^{\prime}, p^{\prime}, \gamma^{\prime}\right\rangle$ be the MDP defined such that 
$S^{\prime}=S \times U, A^{\prime}=A,  
\gamma^{\prime}=\gamma$,
$p^{\prime}\left(\left\langle s^{\prime}, u^{\prime}\right\rangle \mid\langle s, u\rangle, a\right)=\left\{\begin{array}{ll}p\left(s^{\prime} \mid s, a\right) & \text { if } u^{\prime}=\delta_{u}\left(u, \mathcal{F}_L\left(s^{\prime}\right)\right) \\ 
0 & \text { otherwise }\end{array},\right.$
 and $r^{\prime}\left(\langle s, u\rangle, a,\left\langle s^{\prime}, u^{\prime}\right\rangle\right)=\delta_{r}\left(u, u^{\prime}\right)\left(s, a, s^{\prime}\right)$. 
 
 Then any policy for $\mathcal{M}_{\Psi}$ achieves the same expected reward in $\Psi$, and vice versa.
\end{lem}
% \todo{add proof of convergence for \ref{alg:LRM-MaxEntIRL}} solving M_rm should solve the problem.

\subsection{Implementation Details} \label{app:gridworld-implementattion}
\textbf{Generating expert trajectories}
The process flow for generating expert trajectories in the gridworld domain entails first to train an expert policy, $\pi_e$ using a perfect FSA reward structure (Fig.~\ref{fig:app-gridworld-perfect-rm}), then using the expert policy to generate $\mathcal{D}_e$

The final form of the expert demonstrations is represented by  series of state-action transitions, i.e., $(s_1, a_1, s_2, a_2, ..., s_T)$. The actions are represented by single integers from $0$ to $3$. For the low-level state representations, we include both high-level features of the environment (one-hot encoding of one of the high-level positions, e.g., \texttt{\{A, B,...,D, c, m,..\}} etc.) and the low-level positions (one-hot encoding of the 108 grids of the \textit{Office GridWorld} environment).

The following points detail various conditions adhered to while $\mathcal{D}_e$ generation:
\begin{enumerate}
    \item There are K max steps (episode horizon) possible in an episode. 
    \item An episode can end in (t << K) steps if a `done' or `game over' condition is hit (like stepping on obstacles).
    \item Once an optimal trajectory is traversed for 1 round trip (e.g. \textit{Task 4 -- `Patrol ABCD'}: $u_0 \rightarrow u_1 \rightarrow u_2 \rightarrow u_3$), the agent receives a reward of 1. 
    \item After that, the agent takes k random steps.
    \item If not terminated in k steps, \texttt{get\_optimal\_action()} is invoked (from whichever random position the agent is in, creating another successful reward trip completion.
    \item The above step is repeated until max K steps are reached
\end{enumerate}

%%%%%%%%%%%%%%%%%%%%%%%%%%%%%%%%%%%%%%%%%%%%%%%%%%%%%%%%%%%%%%%%%%%%%%
%%%%% APPENDIX C: %%%%%%%%%%%%%%%%%%%%%%%%%%%
%%%%%%%%%%%%%%%%%%%%%%%%%%%%%%%%%%%%%%%%%%%%%%%%%%%%%%%%%%%%%%%%%%%%%%
\section{Reacher Domain Details} \label{appendix:reacher-domain}
In this section we present pertinent details about the experiments conducted on the continuous MuJoCo Reacher domain (Sec.~\ref{sec:experiments-reacher}).
\subsection{ReacherDelivery-v0: Modified Reacher Domain}
We modify the classic OpenAI Gym's~\cite{brockman2016openai} MuJoCo~\cite{todorov2012mujoco} Reacher environment to our purposes here. The original Reacher environment is a two-jointed robot arm, and the goal is to move the robot's end effector (called \textit{fingertip}) close to a target that is spawned at a random position. The \textit{action space} consists of an \textit{action} $(a, b)$ that represents the torques applied at the hinge joints. The \textit{observation space} consists of the sine, cosine angles of the two arms, coordinates of the target, angular velocities of the arms and a 3D distance vector between the target and the reacher's fingertip. The \textit{reward} consists of two parts: i. \textit{reward\_distance} ($R_d$): a measure of how far the fingertip of the reacher (the unattached end) is from the target, with a more negative value assigned with increasing distance; ii. \textit{reward\_control} ($R_c$): a negative reward for penalising actions that are too large. It is measured as the negative squared Euclidean norm of the action, i.e. as $-\sum{action^2} $. The total reward returned is: $reward = reward\_distance + reward\_control$ ($R = R_d + R_c$).

\paragraph{ReacherDelivery} The following code snippet shows how the MuJoCo asset file was modified to add four target locations as colored balls.

\begin{myverbatim}[title={Excerpt from \texttt{reacher\_delivery.xml} file showing added red ball as a target location}]
<!-- RED -->
<body name="red" pos="0 0 0.01">
	<joint armature="0" axis="1 0 0" damping="0" limited="true" name="red_x" pos="0 0 0" range="-.27 .27" ref="0" stiffness="0" type="slide"/>
	<joint armature="0" axis="0 1 0" damping="0" limited="true" name="red_y" pos="0 0 0" range="-.27 .27" ref="0" stiffness="0" type="slide"/>
	<geom conaffinity="0" contype="0" name="red" pos="0 0 0" rgba="0.9 0. 0. 0.3" size=".02" type="sphere"/>
</body>
\end{myverbatim}

We extended the MuJoCo Reacher class with an environment MDP wrapper and added target goals. Following an agent step in the environment, the state proposition labels are updated using the labelling function. The following code snippet exemplifies this, where a target color location proposition is labeled `true' if the fingertip is within a threshold distance from it.

\begin{myverbatim}[title={The labelling function $\mathcal{F}_L$ for \texttt{ReacherDelivery} domain}]
true_props = []
if dist_red < 0.02:
    true_props.append('r')
if dist_green < 0.02:
    true_props.append('g')
if dist_blue < 0.02:
    true_props.append('b')
if dist_yellow < 0.02:
    true_props.append('y')
if cancel:
    true_props.append('c')
\end{myverbatim}

\subsection{Tasks Reward FSA Structures}
\begin{figure}[h!tb]
     \centering
     \begin{subfigure}[t]{0.38\textwidth}
         \centering
         \includegraphics[width=\linewidth]{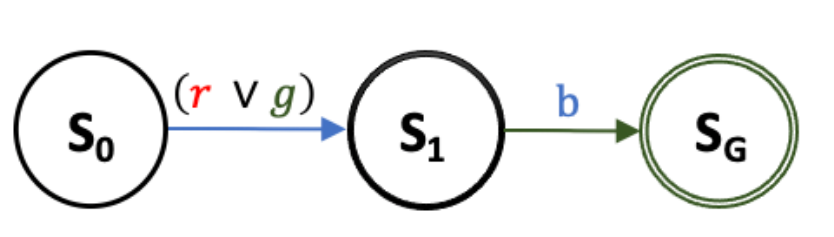}
         \caption{\textbf{OR}: Touch the \textcolor{red}{red} or \textcolor{green}{green} ball, then the \textcolor{blue}{blue} ball} \label{fig:task1-fsa}
     \end{subfigure}
     \hfill
     \begin{subfigure}[t]{0.48\textwidth}
         \centering
         \includegraphics[width=\linewidth]{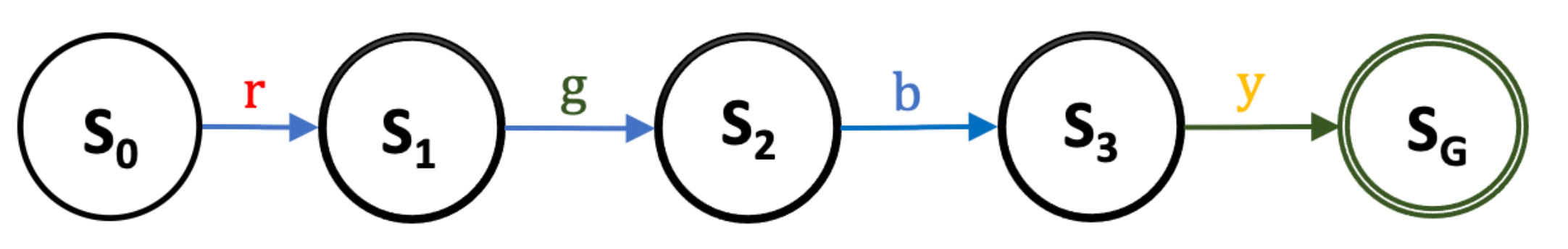}
         \caption{\textbf{Sequential}: Touch the \textcolor{red}{red}, \textcolor{green}{green}, \textcolor{blue}{blue} and \textcolor{yellow}{yellow} ball in order} \label{fig:task2-fsa}
     \end{subfigure}
     \hfill 
     \begin{subfigure}[t]{0.48\textwidth}
         \centering
         \includegraphics[width=\textwidth]{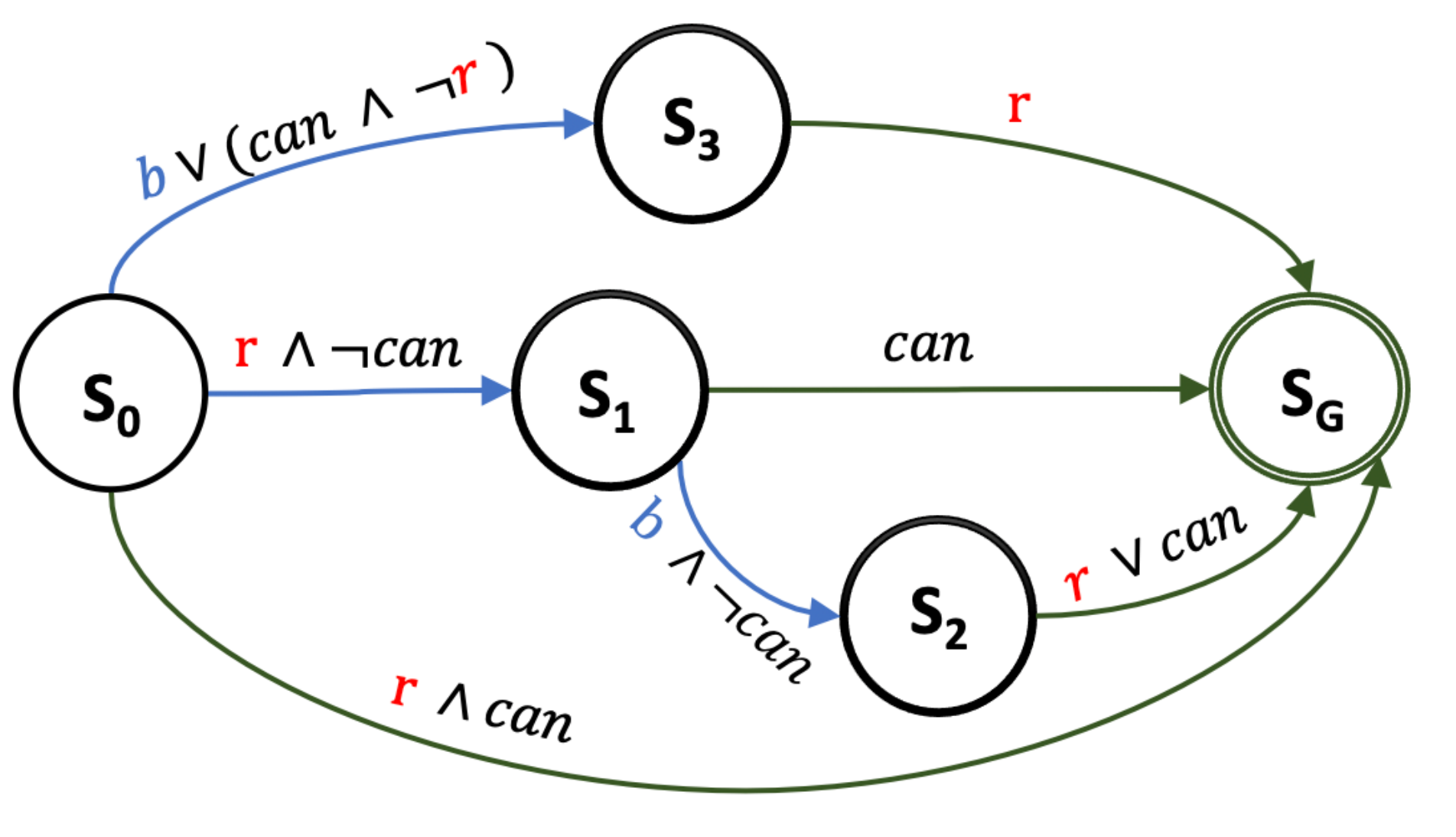}
         \caption{\textbf{IF}: Touch \textcolor{blue}{blue}, unless \textit{cancel}; then touch the \textcolor{red}{red} ball} \label{fig:task3-fsa}
         \label{fig:rm-IF}
     \end{subfigure}
     \hfill
     \begin{subfigure}[t]{0.48\textwidth}
         \centering
         \includegraphics[width=\textwidth]{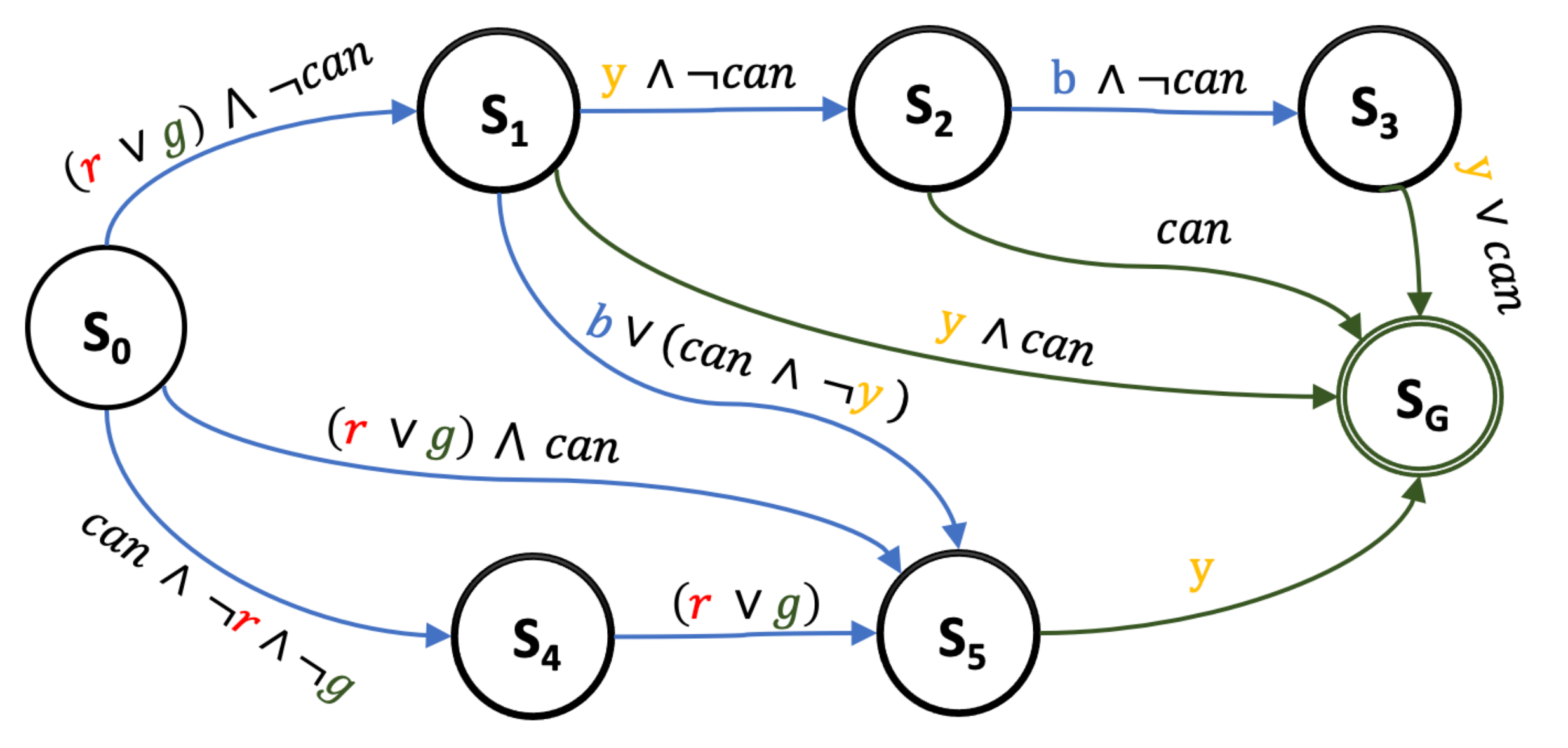}
         \caption{\textbf{Composite}: Touch \textcolor{red}{red} or \textcolor{green}{green}, then \textcolor{blue}{blue}, unless \textit{cancel}, then go to \textcolor{yellow}{yellow}} \label{fig:task4-fsa}
     \end{subfigure}
    \caption{The four increasingly difficult tasks (\texttt{OR, Sequential, IF, Composite}) with the corresponding task descriptions and FSA states}
    \label{fig:app-reacher-fsa-states}
\end{figure}

% Table~\ref{tab:app-reacher-fsa-states} 
Fig.~\ref{fig:app-reacher-fsa-states} shows the (perfect) reward structure for the four tasks along with the task name and the natural language description of the task. The tasks are presented in order of increasing complexity.

% \begin{table}[h!]
% \centering
% \begin{tabular}{ c m{4cm} m{6cm} }
% \toprule
% \textbf{Task} & Description & FSA State Diagram \\
% \midrule
% 	OR & Touch the \textcolor{red}{red} or \textcolor{green}{green} ball, then the \textcolor{blue}{blue} ball
% 	&
% 	\begin{minipage}{.4\textwidth}
%       \includegraphics[width=0.75\linewidth]{./imgs/reacher/fsa-or.pdf}
%     \end{minipage} \\
% \midrule
% %	\raisebox{2.5ex}{Sequential} & Go to the & \includegraphics[width=\linewidth]{imgs/reacher/fsa-sequential.pdf} \\
% 	Sequential & Touch the \textcolor{red}{red}, \textcolor{green}{green}, \textcolor{blue}{blue} and \textcolor{yellow}{yellow} ball in order 
% 	&
% 	\begin{minipage}{0.5\textwidth}
% 		\includegraphics[width=\linewidth]{./imgs/reacher/fsa-sequential.pdf}
% 	\end{minipage} \\ 
% \midrule
% 	IF & Touch \textcolor{blue}{blue}, unless \textit{cancel}; then touch the \textcolor{red}{red} ball
% 	&
% 	\begin{minipage}{.5\textwidth}
%       \includegraphics[width=\linewidth]{./imgs/reacher/fsa-if.pdf}
%     \end{minipage} \\
% \midrule
% 	Composite & Touch \textcolor{red}{red} or \textcolor{green}{green}, then \textcolor{blue}{blue}, unless \textit{cancel}, then go to \textcolor{yellow}{yellow}
% 	&
% 	\begin{minipage}{.5\textwidth}
%       \includegraphics[width=\linewidth]{./imgs/reacher/fsa-composite.pdf}
%     \end{minipage} \\
% \bottomrule
% \end{tabular}
% \vspace{0.75em} 
% \caption{The four increasingly difficult tasks (\texttt{OR, Sequential, IF, Composite}) with the corresponding task descriptions and FSA states}.
% \label{tab:app-reacher-fsa-states}
% \end{table}

\newpage
\subsection{Implementation Details}\label{app:expert-trajectories}
This section outlines the various implementation details for the \textit{ReacherDelivery} domain experiments (Sec.~\ref{sec:experiments-reacher}).

\paragraph{Training Details:} We use SAC as the underlying RL algorithm throughout. The policy network is a tanh squashed Gaussian with mean and standard deviation parameterized by a (64, 64) ReLU MLP with two output heads. The Q-network is a (64,64) ReLU MLP. For optimization we use Adam with learning rate of 0.003 for both the Q-network and policy network. The replay buffer size was 10000 and we used batch size of 256.

For the baselines f-IRL and MaxEntIRL, we used the f-IRL~\cite{f-irl2020} authors' official implementation\footnote{\url{https://github.com/twni2016/f-IRL}}. f-IRL and MaxEntIRL require an estimation of the agent state density. We use kernel density estimation to fit the agent’s density, using Epanechnikov kernel with a bandwidth of 0.2 for pointmass, and a bandwidth of 0.02 for Reacher. At each epoch, we sample 1000 trajectories (30000 states) from the trained SAC to fit the kernel density model.

\begin{figure}[h!tbp]
\centering
  \begin{subfigure}[t]{0.8\textwidth}
    \centering
    \includegraphics[width=\linewidth]{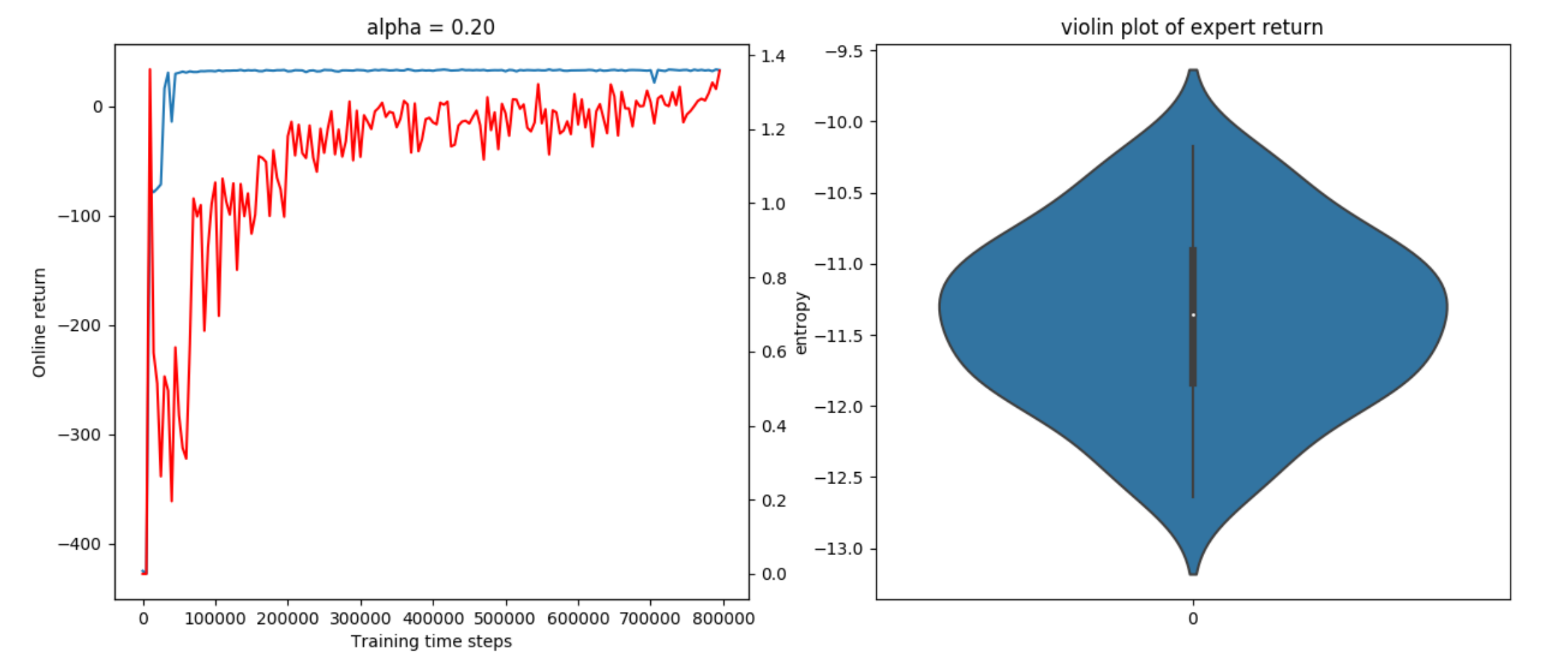}
    \caption{Task: \textbf{Sequential} (nF: 5)}
  \end{subfigure}
  \hfill
  \begin{subfigure}[t]{0.8\textwidth}
    \centering
    \includegraphics[width=\linewidth]{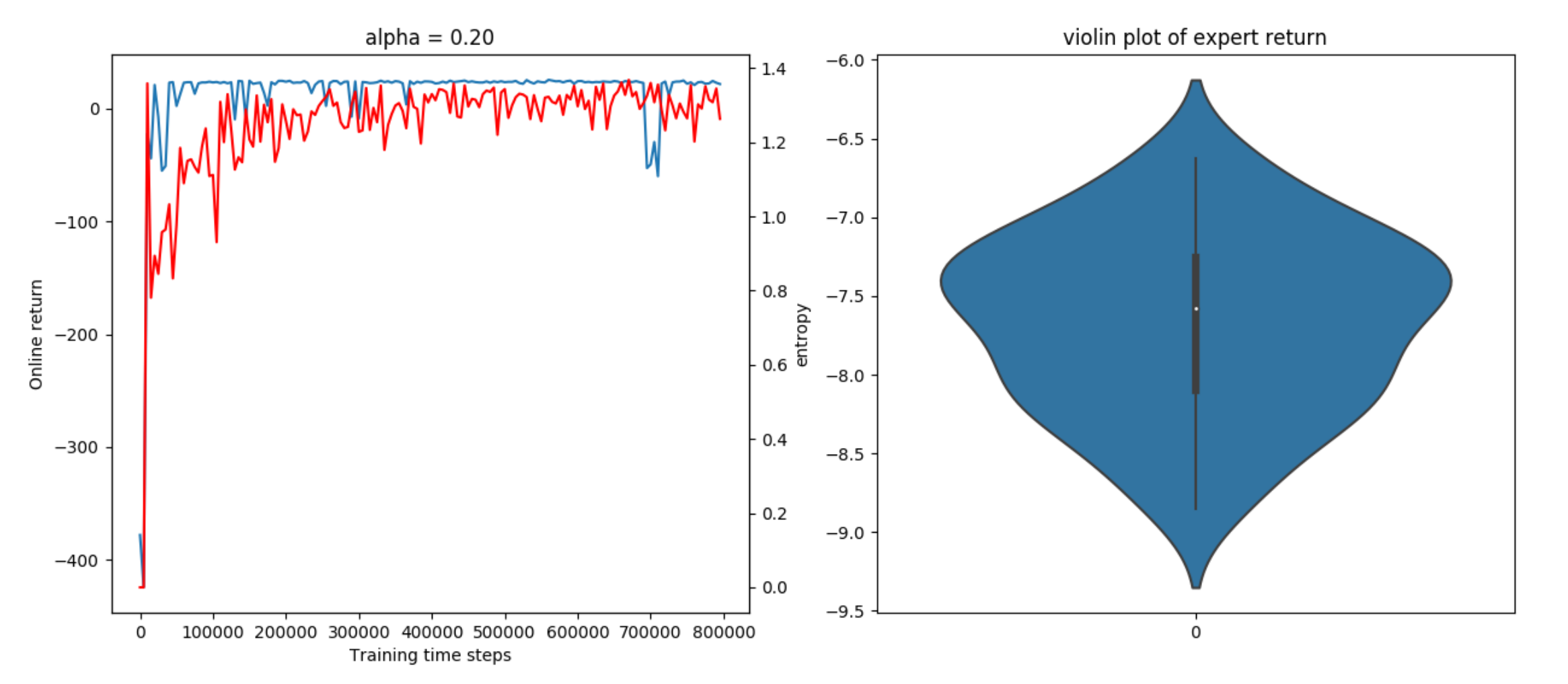}
    \caption{Task: \textbf{IF} (nF: 5)}
  \end{subfigure}
  \medskip
  \begin{subfigure}[t]{0.8\textwidth}
    \centering
    \includegraphics[width=\linewidth]{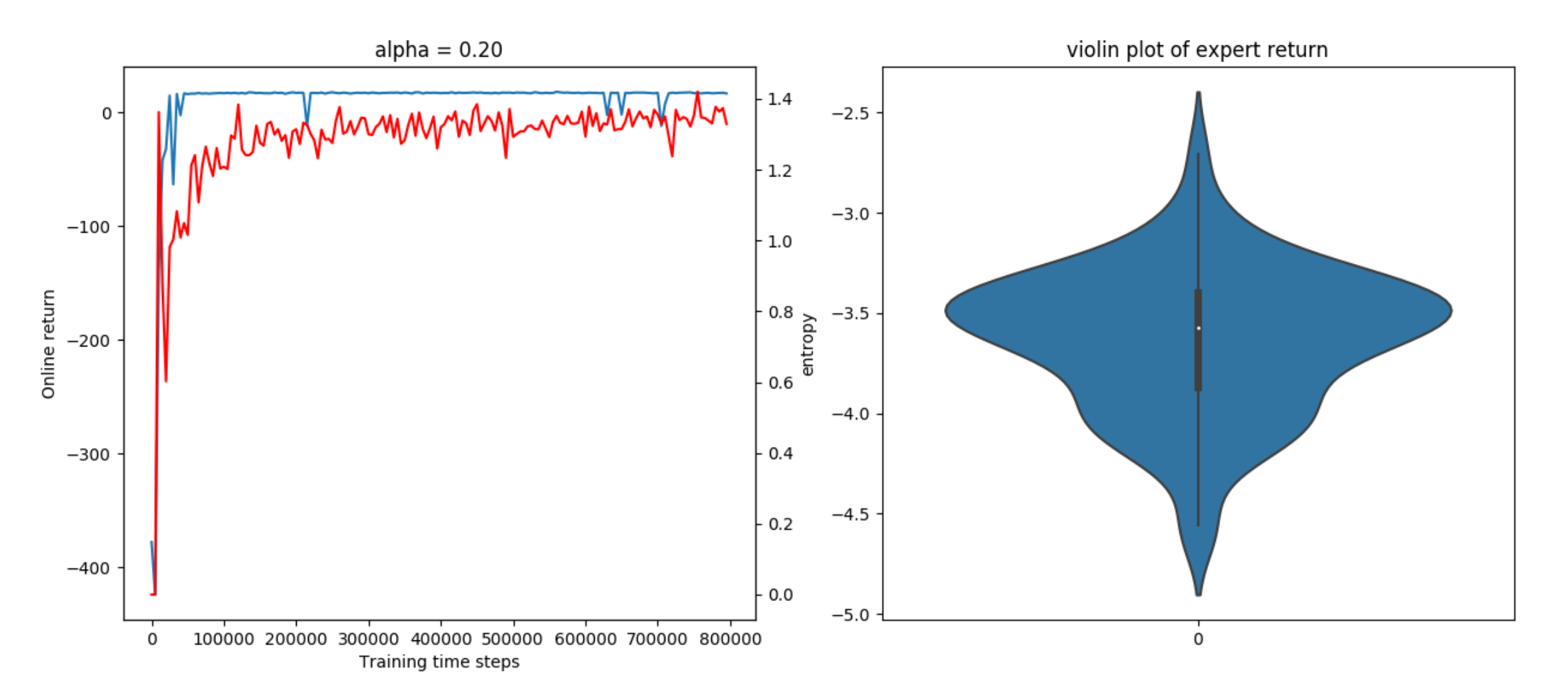}
    \caption{Task: \textbf{OR} (nF: 3)}
  \end{subfigure}
  \hfill
  \begin{subfigure}[t]{0.8\textwidth}
    \centering
    \includegraphics[width=\linewidth]{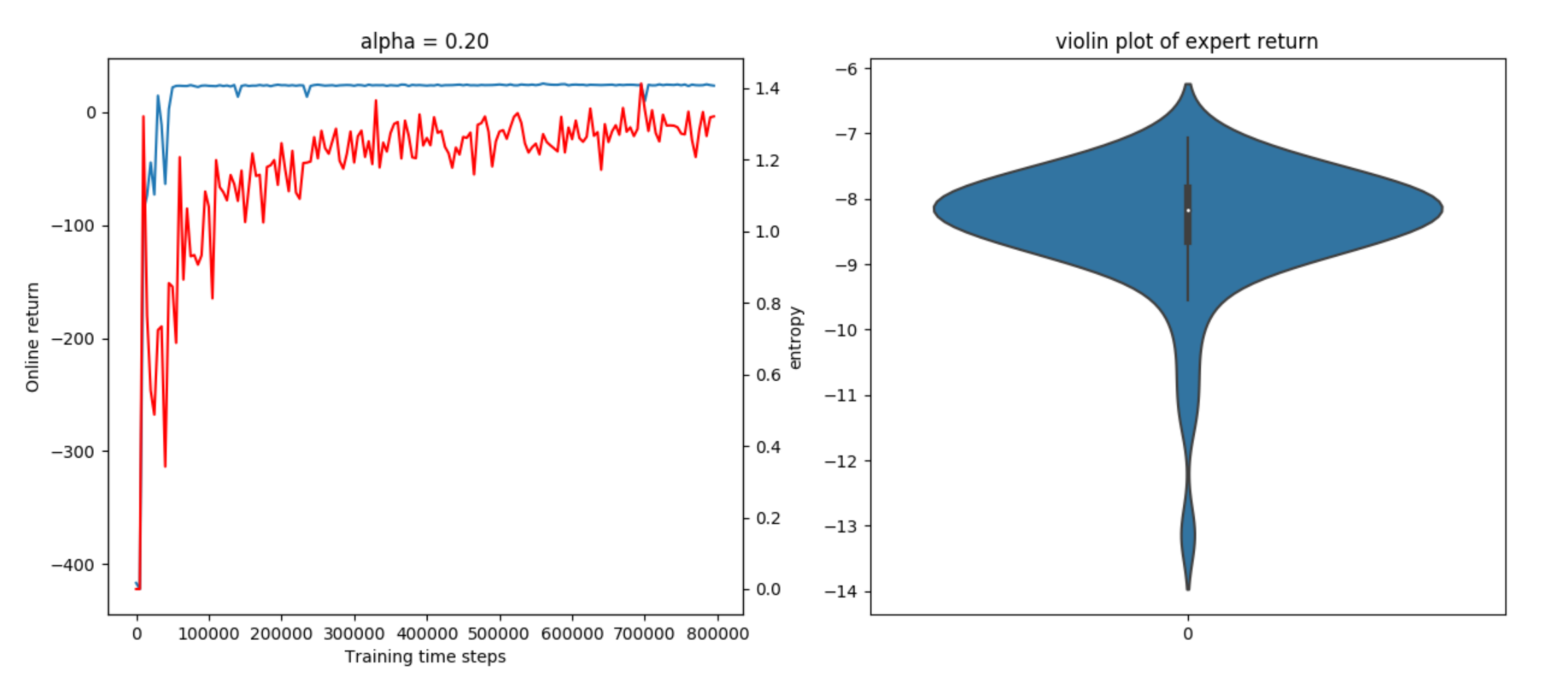}
    \caption{Task: \textbf{Composite} (nF: 7)}
  \end{subfigure}
  \caption{The training curves for expert policy training. The experts were imbued with perfect state based reward structure, where the number of states are given within brackets as `nF' value. \textbf{Left}: The \textcolor{blue}{blue curve} shows reward and the \textcolor{red}{red curve} shows average entropy. \textbf{Right}: The violin plot of expert return density.}
  \label{fig:app:reacher-expert-training}
\end{figure}

\paragraph{Generating Expert Trajectories:} For expert trajectories generation, we first train expert policies imbued with perfect reward structure using SAC for each of the tasks. Fig.~\ref{fig:app:reacher-expert-training} shows the training curve and violin plot expert return density curves of training. 

SAC uses the same policy and critic networks with the learning rate set to 0.003. We train using a batch size of 100, a replay buffer of size 1 million, and set the temperature parameter $\alpha$ to be 0.2. The policy is trained for 1 million timesteps on ReacherDelivery. All algorithms are tested on 16 trajectories collected from the expert stochastic policy.

\paragraph{Evaluation} We compare the trained policies by the baselines (f-IRL< MaxEntIRL) and SMIRL by computing their returns according to the ground truth return on the ReacherDelivery environment. 

\paragraph{Computational Complexity} 
In general, \textit{ceteris paribus}, SMIRL is more sample efficient (i.e. converges to a solution faster) than baseline IRL approaches. Complexity can arise from two factors: i. the domain complexity, say 2D discrete vs. continuous, and ii. The size and complexity of the reward machine (or, FSA) structural motif. We see implications of both in the paper. First, the sample complexity increases with increasing domain complexity (office gridworld vs. Reacher) and this is intuitive. From our experiments, for the second kind of complexity, i.e. with more intricate underlying RM structure, the bottle-neck seems to emanate from the ability to learn the structure. We see that for the hardest (composite) task, the structure was not learned, and increasing sample complexity wouldn't have helped (it would saturate to a suboptimal level). If the structure is learnable, then sample complexity scales with the complexity of the task structure (e.g. 'sequential' vs. 'OR' in Fig. 6).

\section{Baseline objective functions}
\paragraph{f-IRL} We train the three variants of f-IRL: forward KL (fkl), reverse KL (rkl) and Jansen-Shannon (js) that represents the f-divergence metric used by the f-IRL algorithm.

f-divergence~\cite{ali1966general} is a family of distribution divergence metric, which generalizes forward/reverse KL divergence. Formally, let P and Q be two probability distributions over a space $\Omega$, then for a convex and Lipschitz continuous function $f$ such that $f(1) = 0$, the f-divergence of $P$ from $Q$ is defined as:
\begin{equation}
D_{f}(P \| Q):=\int_{\Omega} f\left(\frac{d P}{d Q}\right) d Q
\end{equation}

f-IRL uses state marginal matching by minimizing the f-divergence objective:
\begin{equation}
L_{f}(\theta)=D_{f}\left(\rho_{E}(s) \| \rho_{\theta}(s)\right)
\end{equation}

This objective is realized by computing the exact (analytical) gradient of the preceding f-divergence objective  w.r.t. the reward parameters $\theta$.

% \newpage
\begin{theo}[f-divergence analytic gradient (from~\cite{f-irl2020})] {}
The analytic gradient of the $f$-divergence $L_{f}(\theta)$ between state marginals of the expert $\left(\rho_{E}\right)$ and the soft-optimal agent w.r.t. the reward parameters $\theta$ is given by:
\begin{equation*}
\nabla_{\theta} L_{f}(\theta)=\frac{1}{\alpha T} \operatorname{cov}_{\tau \sim \rho_{\theta}(\tau)}\left(\sum_{t=1}^{T} h_{f}\left(\frac{\rho_{E}\left(s_{t}\right)}{\rho_{\theta}\left(s_{t}\right)}\right), \sum_{t=1}^{T} \nabla_{\theta} r_{\theta}\left(s_{t}\right)\right)
\end{equation*}
where $h_{f}(u) \triangleq f(u)-f^{\prime}(u) u, \rho_{E}(s)$ is the expert state marginal and $\rho_{\theta}(s)$ is the state marginal of the soft-optimal agent under the reward function $r_{\theta}$, and the covariance is taken under the agent's trajectory distribution $\rho_{\theta}(\tau) .^{2}$
\end{theo}

\ifsubfiles
\end{document}
\fi

% \ifsubfiles
%     \subfile{./50-appendix.tex}
% \else
%     \input{50-appendix}
% \fi

\end{document}